\documentclass{article}

\PassOptionsToPackage{numbers,compress}{natbib}
\usepackage[final,main]{neurips_2026}


\usepackage[utf8]{inputenc}
\usepackage[T1]{fontenc}
\usepackage{microtype}
\usepackage{graphicx}
\usepackage{subcaption}
\usepackage{booktabs}
\usepackage{amsmath}
\usepackage{amssymb}
\usepackage{amsfonts}
\usepackage{xcolor}
\usepackage{enumitem}
\usepackage{etoc}
\usepackage[many]{tcolorbox}
\usepackage{wrapfig}
\usepackage{caption}
\usepackage{multirow}
\usepackage{xparse}
\usepackage{tabularx}
\usepackage[hyphens]{url}

\usepackage{hyperref}
\usepackage[capitalize,noabbrev]{cleveref}

\definecolor{figureRed}{HTML}{ff0000}
\definecolor{figureGreen}{HTML}{06b050}
\definecolor{cvprblue}{rgb}{0.21,0.49,0.74}
\hypersetup{
  colorlinks=true,
  linkcolor=cvprblue,
  citecolor=cvprblue,
  urlcolor=cvprblue
}
\colorlet{failgray}{black!60}
\newcommand{\fail}[1]{\textcolor{failgray}{#1}}
\newcommand{\NA}{\textemdash}

\definecolor{skillSize}{HTML}{1B76D2}
\definecolor{skillPosition}{HTML}{483E8C}
\definecolor{skillRotation}{HTML}{C0562A}
\definecolor{skillCount}{HTML}{2E8B57}
\definecolor{skillShape}{HTML}{A0348B}
\definecolor{skillColor}{HTML}{008080}
\newcommand{\SkillSize}{\textcolor{skillSize}{size}}
\newcommand{\SkillPosition}{\textcolor{skillPosition}{position}}
\newcommand{\SkillRotation}{\textcolor{skillRotation}{rotation}}
\newcommand{\SkillCount}{\textcolor{skillCount}{count}}
\newcommand{\SkillShape}{\textcolor{skillShape}{shape}}
\newcommand{\SkillColor}{\textcolor{skillColor}{color}}

\newcommand{\CatA}{\textbf{A}}
\newcommand{\CatB}{\textbf{B}}
\NewDocumentCommand{\CatC}{o g}{%
  \textbf{C}%
  \IfValueT{#1}{~(#1)}%
  \IfValueT{#2}{~(#2)}%
}
\newcommand{\CatD}{\textbf{D}}

\newtcolorbox{findingbox}{
  colback=black!2,
  colframe=black,
  boxrule=0.5pt,
  arc=2pt,
  left=4pt,right=4pt,top=3pt,bottom=3pt
}

\title{Do Diffusion Models Learn to Generalize\\Basic Visual Skills?}

\author{%
  Amish Sethi \quad Boya Zeng \quad Wenhao Chai \quad Zhuang Liu \\
  Princeton University
}

\begin{document}

\maketitle

\begin{abstract}
Diffusion models generate high-quality images, but how well they generalize is hard to measure.
Current evaluations rely on human preference and prompt-following benchmarks for text-to-image models and on distribution-level metrics such as Fr\'echet Inception Distance (FID) and Inception Score (IS) for class-conditional models, but these cannot tell whether models learn basic visual skills or largely match training patterns.
To address this gap, we introduce the Visual Generative Lab (VGL), a controlled experimental framework for studying how diffusion models learn and generalize basic visual skills, including \SkillSize, \SkillPosition, \SkillRotation, \SkillCount, \SkillColor, \SkillShape, and their compositions.
By generating synthetic training data for each skill from explicit rules, VGL allows precise measurement of generalization with task-specific, rule-based metrics.
Training models from scratch on these data reveals three patterns in this controlled setting:
(1) with our default design, extrapolation is limited: \SkillSize\ and \SkillPosition\ degrade within a few pixels of the training range, \SkillCount\ reaches only about $50\%$ accuracy on out-of-range counts, and \SkillRotation\ extrapolation accuracy varies widely across seeds;
(2) for compositional generalization, the number of distinct basic visual skill combinations matters more than dataset size;
(3) skills mostly fail independently, except \SkillPosition.
Code and data-generation scripts: \url{https://github.com/AmishSethi/visual-generative-lab}.

\end{abstract}

\section{Introduction}
\label{sec:intro}

\suppressfloats[t]
\begin{figure*}[t]
\centering
\includegraphics[width=0.94\textwidth]{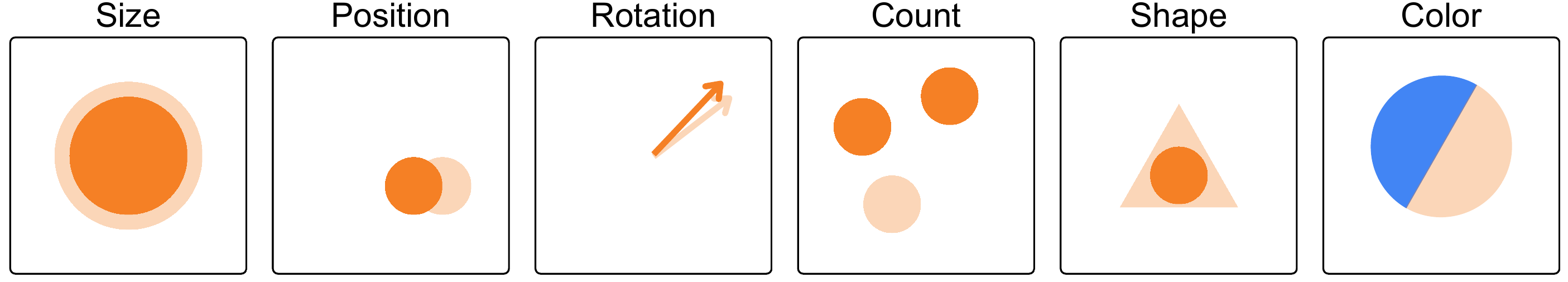}
\caption{
\textbf{The Visual Generative Lab (VGL).}
VGL trains diffusion models from scratch on synthetic renders labeled with explicit skill values.
Each panel varies one skill while holding the others fixed.
Models are trained on a subset of skill values or combinations and evaluated on unseen ones within or beyond the training range.
}
\label{fig:teaser}
\end{figure*}
\suppressfloats[t]

The rise of diffusion models~\citep{ho2020denoising,song2021score,nichol2021improved,dhariwal2021diffusion} has demonstrated strong capabilities in generating high-quality images, from class-conditional synthesis~\citep{peebles2023scalable,hoogeboom2023simple} to text-to-image generation~\citep{nichol2022glide,rombach2022high,saharia2022photorealistic,ramesh2022hierarchical}.
These advances are driven by architectural ideas such as the U-Net~\citep{ronneberger2015u} and transformer-based designs~\citep{peebles2023scalable,bao2023all}, and by conditioning mechanisms like cross-attention~\citep{rombach2022high} and adaptive normalization~\citep{perez2018film,peebles2023scalable}.
Recent work explores synthetic datasets~\citep{johnson2017clevr,yi2020clevrer} for controlled empirical analysis~\citep{okawa2023compositional,park2024emergence} and theoretical analysis~\citep{hirono2024understanding} of diffusion behavior, yet comprehensive controlled studies of how diffusion models learn and compose basic visual skills remain limited.

Existing evaluations of generative models fall short in two ways.
First, we lack granular insight into how diffusion models fail.
Class-conditional models on ImageNet~\citep{deng2009imagenet} rely on distribution-level metrics like Fr\'echet Inception Distance (FID) and Inception Score (IS)~\citep{salimans2016improved,heusel2017gans,kynkaanniemi2019improved,naeem2020reliable,dhariwal2021diffusion,peebles2023scalable}.
These metrics reveal neither which basic visual skills models struggle with nor how they fail.
Large text-to-image systems are judged by human preference~\citep{jiang2024genai} and by prompt-following benchmarks like GenEval~\citep{ghosh2023geneval}, T2I-CompBench~\citep{huang2023t2i}, DPG-Bench~\citep{hu2024ella}, and OneIG-Bench~\citep{chang2025oneig}, which decompose evaluation into skills such as counting, spatial relations, and attribute binding~\citep{lee2023holistic,bakr2023hrs,hu2023tifa}.
However, these benchmarks evaluate only pre-trained models whose behavior is shaped by hidden training distributions, and their scores cannot separate generalization from retrieval of familiar patterns~\citep{zhang2017understanding,arpit2017closer,somepalli2023diffusion,carlini2023extracting,jain2025classifier}.

Second, we lack control over training distributions.
ImageNet and large-scale image-text corpora are fixed collections where multiple visual factors vary simultaneously~\citep{deng2009imagenet,desai2021redcaps}.
This makes it difficult to isolate which \emph{skills} are generalized and to study how conditioning formats, data coverage, or model design affect the learning of basic visual skills.
These gaps motivate an evaluation setting that measures whether models generalize specific basic visual skills beyond the support of their training data.

We introduce the Visual Generative Lab (VGL), a framework for training diffusion models from scratch on synthetic geometric data with explicitly controlled training distributions.
Conceptually, VGL is inspired by computations in the human ventral visual stream, which build object-centered representations tolerant to variation in \SkillPosition, \SkillSize, and viewpoint (\SkillRotation)~\citep{marr1982vision,DiCarlo2012}.
In VGL, we choose exactly which skill values, and which combinations of values, appear in training.
Because skill values are provided as explicit numeric or categorical conditions, extrapolation in VGL is a clean distribution-shift test in the conditioning space.
Such extrapolation reveals whether a model captures the underlying structure of a skill or falls back to the nearest training value (nearest-support behavior).
We then conduct systematic experiments that reveal three patterns about how diffusion models learn and generalize basic visual skills.

\paragraph{Extrapolation is limited.}
We evaluate single-skill generalization by training models on a restricted set of skill values and testing on held-out values within and beyond the training range.
With our default design, \SkillSize\ and \SkillPosition\ degrade within a few pixels of the training range, and \SkillCount\ reaches only $52.5\%$ on out-of-range counts.
This pattern suggests that, with current diffusion models, most basic visual skills require explicit data coverage (Section~\ref{sec:single_property}).

\paragraph{The number of distinct combinations matters more than dataset size.}
When combining multiple basic visual skills, e.g., generating objects with a specific \SkillColor\ and \SkillShape, we systematically vary both the percentage of combinations seen during training (\emph{coverage}) and the number of samples per combination (\emph{density}).
Models trained on high coverage with few samples per combination outperform models trained on low coverage with many samples per combination (Section~\ref{sec:compositional}).

\paragraph{Skills mostly fail independently, except \SkillPosition.}
We test models conditioned jointly on two to four skills by moving one requested value out of distribution (OOD) while keeping the others in range. We find that the in-range skills mostly keep their accuracy, except when \SkillPosition\ is OOD, which can also lower in-range \SkillSize, \SkillRotation, and \SkillShape\ (Section~\ref{sec:distribution_learning}).

Together, these results suggest that, at least in our controlled setting, diffusion models do not yet generalize basic visual skills far beyond what their training data covers.
If this holds more broadly for large-scale visual generative models, data providers and model trainers may need to actively curate data to cover the skill values and combinations users ask for rather than simply relying on more data alone, and evaluations should test models outside that coverage before crediting them with generalization.

\section{Visual Generative Lab (VGL)}
\label{sec:experimental_setup}

\begin{table*}[t]
\centering
\small
\setlength{\tabcolsep}{6pt}
\begin{tabular}{@{}llll@{}}
\toprule
\textbf{Skill} & \textbf{Type} & \textbf{Domain} & \textbf{Extrapolation values} \\
\midrule
\SkillSize & continuous & radius $\in [5, 20]$ px & $\{1,2,3,4\} \cup \{21,22,\dots,30\}$ px \\
\SkillPosition & continuous (2D) & $(x,y) \in [-18, 18]^2$ px & $L_\infty \geq 4$ px outside $[-18,18]^2$ (320 positions) \\
\SkillRotation & continuous & angle $\in [45^\circ, 315^\circ]$ & $\{0^\circ, \dots, 44^\circ\} \cup \{316^\circ, \dots, 359^\circ\}$ \\
\SkillCount & discrete & $\{2, 3, \dots, 7\}$ objects & $\{0, 1, 8, 9\}$ \\
\SkillShape & categorical & 4 classes & \NA \\
\SkillColor & categorical & 8 classes & \NA \\
\bottomrule
\end{tabular}
\vspace{0.3em}
\caption{\textbf{Training domains and extrapolation values for VGL skills.}
\SkillSize, \SkillPosition, and \SkillRotation\ also have an interpolation split of unseen values inside the domain. \SkillShape\ and \SkillColor\ are evaluated on all their classes. Appendix~\ref{appendix:datasets} gives the full value grids and splits.}
\label{tab:split_values}
\end{table*}

\paragraph{Motivation.}
Human visual intelligence builds object representations that are tolerant to changes in \SkillPosition, \SkillSize, and viewpoint (\SkillRotation) via computations in the ventral visual stream~\citep{marr1982vision,DiCarlo2012}.
Complex scene understanding and generation build on these skills and their combinations.
To test whether diffusion models learn such skills or largely match their training data, we need data that isolate one skill at a time and whose distribution we set exactly.
Real-world datasets like ImageNet vary many factors at once, so it is hard to trace model behavior to a specific design choice or property of the data.

\paragraph{Defining generalization.}
We define generalization in VGL relative to the skill values that appear in training.
We use \SkillSize\ as a running example (Figure~\ref{fig:teaser} shows all six skills).
We parameterize object size by radius $r \in [r_{\min}, r_{\max}]$ and render circles on a $64 \times 64$ pixel canvas.
During training, we control which radius values appear in the training set, e.g., the 16 integer radii between 5 and 20 pixels.
The model receives the target skill value as a conditioning signal during training and inference.
At test time, we evaluate three forms of generalization, namely (1) interpolation on intermediate values not seen during training, such as radius 12.5; (2) extrapolation on values beyond the training range, such as radius 25; and (3) composition with other basic visual skills (Section~\ref{sec:compositional}).

\paragraph{Conditioning format.}
In VGL, each image is associated with explicit skill values.
Continuous skills and \SkillCount\ are provided as numeric values, and categorical skills are provided as discrete labels.
Our default transformer backbone conditions on these skills by embedding each value into a token and prepending the skill tokens to the image patch tokens, so the denoiser processes conditions and image content jointly through self-attention.
Appendix~\ref{appendix:architecture} specifies the exact embedding and tokenization used for each skill.

\paragraph{Evaluation metrics.}
VGL relies on skill-specific evaluation protocols that measure each skill directly in the generated image with simple computer vision operations, without relying on perceptual metrics~\citep{heusel2017gans,kynkaanniemi2019improved,naeem2020reliable}.
For \SkillSize, we apply Otsu thresholding~\citep{otsu1979threshold} and morphological cleaning to obtain a binary mask, estimate the radius from the mask area as $r_{\text{pred}} = \sqrt{\text{area}/\pi}$, and consider a generation correct when the intersection-over-union (IoU) between predicted and ground-truth masks is at least 0.90.
We measure the other skills with the mask centroid (\SkillPosition), template matching (\SkillRotation), connected components (\SkillCount), template-based shape classification (\SkillShape), and nearest-neighbor color assignment (\SkillColor).
We also report errors such as mean absolute error (MAE).
Appendix~\ref{appendix:eval_protocol} gives the full metric implementations, and Figure~\ref{fig:qualitative} and Appendix~\ref{appendix:qualitative_analysis} show qualitative examples.

\section{Single-Skill Generalization}
\label{sec:single_property}

We use VGL to study the generalization capability of diffusion models on four basic visual skills, namely object \SkillSize, \SkillPosition, \SkillRotation, and \SkillCount.
We also evaluate how common modeling design choices affect generalization on each skill.

\begin{table*}[t]
\centering
\scriptsize
\setlength{\tabcolsep}{2.5pt}
\resizebox{\textwidth}{!}{%
\begin{tabular}{@{}lcccccccccccc@{}}
\toprule
 & \multicolumn{3}{c}{\SkillSize} & \multicolumn{3}{c}{\SkillPosition} & \multicolumn{3}{c}{\SkillRotation} & \multicolumn{3}{c}{\SkillCount} \\
\cmidrule(lr){2-4} \cmidrule(lr){5-7} \cmidrule(lr){8-10} \cmidrule(lr){11-13}
\textbf{Model} &
\textbf{Train} & \textbf{Interp.} & \textbf{Extrap.} &
\textbf{Train} & \textbf{Interp.} & \textbf{Extrap.} &
\textbf{Train} & \textbf{Interp.} & \textbf{Extrap.} &
\textbf{Train} & \textbf{Interp.} & \textbf{Extrap.} \\
\midrule
DiT-S/2 (baseline) & $100.0_{\pm 0.0}$ & $96.8_{\pm 0.7}$ & $51.5_{\pm 7.5}$ & $100.0_{\pm 0.0}$ & $97.4_{\pm 3.2}$ & $29.7_{\pm 16.8}$ & $100.0_{\pm 0.0}$ & $100.0_{\pm 0.0}$ & $68.0_{\pm 28.9}$ & $100.0_{\pm 0.0}$ & \textemdash{} & $52.5_{\pm 34.5}$ \\
\midrule
+ sinusoidal emb. & $100.0_{\pm 0.0}$ & $96.9_{\pm 0.4}$ & $7.1_{\pm 0.0}$ & $100.0_{\pm 0.0}$ & $100.0_{\pm 0.0}$ & $0.0_{\pm 0.0}$ & $100.0_{\pm 0.0}$ & $100.0_{\pm 0.0}$ & $26.6_{\pm 1.6}$ & $98.6_{\pm 1.3}$ & \textemdash{} & $20.4_{\pm 11.8}$ \\
+ rotary emb. & $100.0_{\pm 0.0}$ & $95.3_{\pm 1.5}$ & $9.9_{\pm 3.9}$ & $100.0_{\pm 0.0}$ & $100.0_{\pm 0.0}$ & $0.0_{\pm 0.0}$ & $100.0_{\pm 0.0}$ & $100.0_{\pm 0.0}$ & $38.1_{\pm 6.2}$ & $100.0_{\pm 0.0}$ & \textemdash{} & $36.3_{\pm 1.3}$ \\
+ \textbf{linear emb.} & $100.0_{\pm 0.0}$ & $96.8_{\pm 0.7}$ & $51.5_{\pm 7.5}$ & $100.0_{\pm 0.0}$ & $97.4_{\pm 3.2}$ & $29.7_{\pm 16.8}$ & $100.0_{\pm 0.0}$ & $100.0_{\pm 0.0}$ & $68.0_{\pm 28.9}$ & $100.0_{\pm 0.0}$ & \textemdash{} & $52.5_{\pm 34.5}$ \\
\midrule
+ AdaLN & $100.0_{\pm 0.0}$ & $85.9_{\pm 8.7}$ & $31.0_{\pm 4.2}$ & $100.0_{\pm 0.0}$ & $100.0_{\pm 0.0}$ & $61.5_{\pm 14.4}$ & $100.0_{\pm 0.0}$ & $100.0_{\pm 0.0}$ & $58.5_{\pm 11.8}$ & $100.0_{\pm 0.0}$ & \textemdash{} & $72.5_{\pm 3.3}$ \\
+ \textbf{concatenation} & $100.0_{\pm 0.0}$ & $96.8_{\pm 0.7}$ & $51.5_{\pm 7.5}$ & $100.0_{\pm 0.0}$ & $97.4_{\pm 3.2}$ & $29.7_{\pm 16.8}$ & $100.0_{\pm 0.0}$ & $100.0_{\pm 0.0}$ & $68.0_{\pm 28.9}$ & $100.0_{\pm 0.0}$ & \textemdash{} & $52.5_{\pm 34.5}$ \\
\midrule
+ VAE latent & $99.1_{\pm 0.3}$ & $95.9_{\pm 0.7}$ & $7.1_{\pm 0.0}$ & $100.0_{\pm 0.0}$ & $100.0_{\pm 0.0}$ & $0.0_{\pm 0.0}$ & $100.0_{\pm 0.0}$ & $100.0_{\pm 0.0}$ & $21.2_{\pm 6.2}$ & $93.3_{\pm 0.8}$ & \textemdash{} & $48.8_{\pm 15.2}$ \\
+ \textbf{pixel} & $100.0_{\pm 0.0}$ & $96.8_{\pm 0.7}$ & $51.5_{\pm 7.5}$ & $100.0_{\pm 0.0}$ & $97.4_{\pm 3.2}$ & $29.7_{\pm 16.8}$ & $100.0_{\pm 0.0}$ & $100.0_{\pm 0.0}$ & $68.0_{\pm 28.9}$ & $100.0_{\pm 0.0}$ & \textemdash{} & $52.5_{\pm 34.5}$ \\
\midrule
+ flow matching & $100.0_{\pm 0.0}$ & $96.7_{\pm 0.3}$ & $47.3_{\pm 3.9}$ & $100.0_{\pm 0.0}$ & $98.9_{\pm 1.9}$ & $27.8_{\pm 26.2}$ & $100.0_{\pm 0.0}$ & $100.0_{\pm 0.0}$ & $51.0_{\pm 6.1}$ & $100.0_{\pm 0.0}$ & \textemdash{} & $24.2_{\pm 1.4}$ \\
+ U-Net (matched \#params) & $100.0_{\pm 0.0}$ & $67.8_{\pm 1.4}$ & $21.2_{\pm 5.8}$ & $100.0_{\pm 0.0}$ & $100.0_{\pm 0.0}$ & $42.7_{\pm 36.4}$ & $100.0_{\pm 0.0}$ & $100.0_{\pm 0.0}$ & $11.2_{\pm 0.1}$ & $99.7_{\pm 0.5}$ & \textemdash{} & $79.6_{\pm 4.7}$ \\
\midrule
+ DiT-L (higher \#params) & $100.0_{\pm 0.0}$ & $85.3_{\pm 4.1}$ & $44.5_{\pm 1.1}$ & $100.0_{\pm 0.0}$ & $98.7_{\pm 1.1}$ & $42.5_{\pm 20.4}$ & $100.0_{\pm 0.0}$ & $100.0_{\pm 0.0}$ & $62.0_{\pm 15.6}$ & $100.0_{\pm 0.0}$ & \textemdash{} & $72.1_{\pm 22.0}$ \\
\bottomrule
\end{tabular}%
}
\vspace{0.3em}
\caption{
\textbf{Single-skill generalization under different design choices.}
Entries are mean$\pm$std accuracy (\%) over three seeds, or ten for the \SkillRotation\ baseline.
Success thresholds are IoU $\geq 0.90$ for \SkillSize, 2~px for \SkillPosition, $5^\circ$ for \SkillRotation, and an exact match for \SkillCount.
\SkillCount\ has no interpolation split.
Each row below the baseline changes one design choice. \textbf{Bold} rows are the defaults we adopt, so they repeat the baseline.
}
\label{tab:single_property_results}
\end{table*}

\subsection{Generalization}

\begin{findingbox}
\textbf{Finding 1.} Diffusion models extrapolate only slightly beyond the training support.
\end{findingbox}

We first train a baseline diffusion model on each skill and analyze how it generalizes.
Table~\ref{tab:single_property_results} reports its accuracy, and Table~\ref{tab:split_values} lists the training domains and extrapolation values.
The \emph{train} columns denote in-support evaluation on newly generated samples whose skill values appear in the training set.
Results are means over three random seeds, or ten for the \SkillRotation\ baseline.

\paragraph{Setup.}
By default, we use the DiT-S/2~\citep{peebles2023scalable} architecture and train with the standard DDPM noise prediction objective~\citep{ho2020denoising}.
Models are trained for 1000 epochs using AdamW~\citep{loshchilov2019decoupled} with learning rate $1{\times}10^{-4}$, except \SkillCount, which needs 3000 epochs to fit its training values (details in Appendix~\ref{appendix:count_geometry}).
At evaluation, we sample with 250-step ancestral DDPM sampling using EMA weights.
By default, we do not use classifier-free guidance~\citep{ho2022classifier}, i.e., guidance scale 1.
In Appendix~\ref{appendix:cfg_size_ablation}, we show that adding guidance does not remove the \SkillSize\ extrapolation limit.
Unless otherwise stated, we generate 20 samples per condition for the accuracy metrics.

\begin{figure*}[!htb]
\centering
\includegraphics[width=0.85\textwidth]{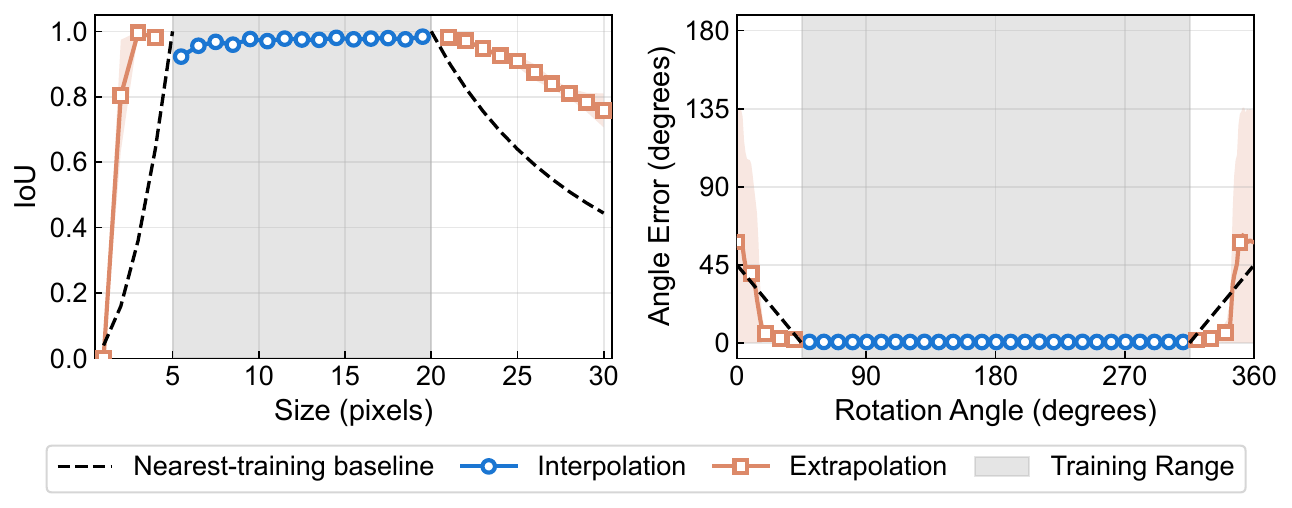}
\caption{\textbf{\SkillSize\ interpolates well but extrapolates only slightly, and \SkillRotation\ extrapolation accuracy varies widely across seeds.}
Left. IoU per radius. Right. Angular error per angle. Lines are means over the seeds of Table~\ref{tab:single_property_results} (three for \SkillSize, ten for \SkillRotation), and bands are $\pm 1$ standard deviation over seeds. The dashed line is a baseline that outputs the nearest training value (Appendix~\ref{appendix:eval_protocol}).
The \SkillRotation\ band is wide near $0^\circ$, the angle farthest from the training range, because the seeds differ: four keep the error below $5^\circ$ at every angle, and the others make large errors. Figure~\ref{fig:radius_rotation_seeds} shows every seed.}
\label{fig:radius_rotation}
\end{figure*}

\begin{wrapfigure}{R}{0.5\textwidth}
    \centering
    \includegraphics[width=0.45\textwidth]{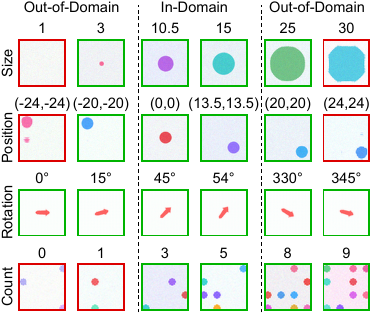}
    \captionsetup{width=0.45\textwidth}\caption{\textbf{Generated samples for each skill.} \textcolor{figureGreen}{Green}/\textcolor{figureRed}{red} borders mark success/failure according to the per-skill criteria.}
    \label{fig:qualitative}
    \vspace{-0.8em}
\end{wrapfigure}

\paragraph{\SkillSize\ and \SkillPosition.}
Prior work has shown that generative models often degrade when asked to generate objects of unseen sizes or positions~\citep{choi2021toward}, and we revisit this question in VGL where the training and test distributions are specified directly.
For \SkillSize, we train on circles with 16 integer radii between 5 and 20 pixels and evaluate extrapolation on radii in $\{1, 2, 3, 4\}$ and $\{21, 22, \dots, 30\}$ pixels.
Training radii reach $100\%$ accuracy and interpolation reaches $96.8\%$, while extrapolation drops to $51.5\pm 7.5\%$.
Above the training range, every seed succeeds (at least half of the samples pass) through radius 24 and fails from 27, while below it, every seed succeeds at radius 3 and fails at radius 1. The extrapolation ability is therefore about 4 px above and 2 px below the training range. The model beats the nearest-training baseline (dashed line in Figure~\ref{fig:radius_rotation}) at every out-of-range radius except 1, even where it fails (examples in Figure~\ref{fig:qualitative}).

For \SkillPosition, we train on circles at $32 \times 32$ uniformly spaced positions in $[-18, 18]^2$ pixels and evaluate the far-OOD frame of positions whose $L_\infty$ distance to the training square is at least 4 pixels (320 positions).
Training and interpolated positions reach $100\%$ and $97.4\pm 3.2\%$ accuracy at the 2-pixel tolerance, while far-OOD extrapolation drops to $29.7\pm 16.8\%$, with accuracy going from $49\%$ at 4.6 px outside training to $11\%$ at 5.8 px out.
Figure~\ref{fig:qualitative} shows that the model can generate multiple circles or otherwise break the intended single-object structure under out-of-range position queries.

Across both skills, the model produces correct outputs on training and interpolated values but extrapolates only a small amount before performance degrades.
To test whether more training data closes the extrapolation gap, we train the baseline on nested sets of 10k to 80k images for each of the four skills at a budget of 78k to 80k optimizer steps.
Training and interpolation accuracy stay flat, and extrapolation moves within noise on every skill (Appendix~\ref{appendix:data_scaling}).

\paragraph{\SkillRotation.}
We next ask whether diffusion models can learn precise object rotations.
We render asymmetric arrows and train on every degree between $45^\circ$ and $315^\circ$, then evaluate angles outside this interval.
Training and interpolation reach $0.30^\circ$ and $0.54^\circ$ mean angular error, and the ten-seed mean extrapolation accuracy is $68.0\%$ at the $5^\circ$ threshold.
On the two seeds with the lowest extrapolation MAE (seeds 4 and 6 in Table~\ref{tab:rotation_per_seed_main}, $0.55^\circ$ and $0.48^\circ$), error stays near zero at every out-of-range angle, far below the nearest-training baseline (Figure~\ref{fig:radius_rotation_seeds}).
\SkillRotation\ is the only skill for which some seeds reach $100\%$ accuracy on the entire out-of-range set.
Its input $[\sin\theta, \cos\theta]$ is periodic, so $\theta$ and $\theta + 360^\circ$ share one embedding (Appendix~\ref{appendix:architecture}) and every out-of-range angle lies within $45^\circ$ of the training range.
We do note that relative to the width of each training range, the out-of-range sets of \SkillSize\ and \SkillCount\ reach farther (Appendix~\ref{appendix:matched_distance}).

The ten seeds fall into three different extrapolation behaviors.
Four seeds extrapolate across the whole out-of-range set with mean angular error below $2^\circ$, three seeds are middling with errors of $8^\circ$ to $15^\circ$, and three seeds collapse near $0^\circ$, the angle farthest from the training range, with errors above $30^\circ$.
Appendix~\ref{appendix:rotation_ablations} provides per-seed numbers, coverage ablations, and qualitative examples near $0^\circ$.

\paragraph{\SkillCount.}
Recent studies show that pre-trained text-to-image diffusion models often fail to generate the correct number of objects~\citep{guo2025text,kang2025counting}.
We train on images containing 2 to 7 circles and evaluate extrapolation to counts 0, 1, 8, and 9.
The baseline model is perfect on train counts, but reaches only $52.5\%$ on counts just outside the training range.
The model shown in Figure~\ref{fig:qualitative} draws one to three objects when asked for 0 and usually succeeds on 1, 8, and 9, while one of the three seeds in Table~\ref{tab:single_property_results} fails on every out-of-range count except 0.
Appendix~\ref{appendix:qualitative_analysis} shows more examples of these failures.

To check whether these trends depend on our default setting, we repeat the single-skill experiments on textured renders (Appendix~\ref{appendix:realistic}) and at $128\times128$ resolution (Appendix~\ref{appendix:hires}).
On textured renders, \SkillRotation\ extrapolates best and \SkillCount\ worst.
At $128\times128$, \SkillSize\ extrapolation is similar and \SkillRotation\ again splits across seeds.
We also prompt eight pre-trained text-to-image models for the same four skills (Appendix~\ref{appendix:t2i}).
None of them follows the requested \SkillSize, and the best reaches $58\%$ on \SkillCount.

\subsection{Design choices}

The baseline extrapolates \SkillSize, \SkillPosition, and \SkillCount\ only slightly, and its \SkillRotation\ extrapolation varies widely across seeds.
These trends may depend on modeling choices.
We therefore change one design choice at a time and examine how conditioning, data representation, training objective, architecture, and model size affect generalization on the four basic visual skills.
The baseline already includes the choices we adopt (the \textbf{bold} rows in Table~\ref{tab:single_property_results}).

\paragraph{Conditioning embedding.}
We compare three ways of embedding condition information: sinusoidal~\citep{vaswani2017attention}, rotary~\citep{su2024roformer}, and a simple linear projection.
Sinusoidal and rotary embeddings are standard for encoding token positions in transformers, and we adapt them for scalar conditions.
The linear embedding instead applies a learnable layer to the scalar value, or to $[\sin\theta, \cos\theta]$~\citep{zhou2019continuity} for \SkillRotation\ (Appendix~\ref{appendix:architecture}).
As shown in Table~\ref{tab:single_property_results}, it outperforms the other two on \SkillSize, \SkillPosition, and \SkillRotation\ extrapolation and has the highest mean on \SkillCount.
For example, \SkillSize\ extrapolation reaches $51.5\%$ versus $7.1\%$ and $9.9\%$, and \SkillRotation\ extrapolation reaches $68.0\%$ versus $26.6\%$ and $38.1\%$.
This suggests that embeddings that preserve the structure of the condition (monotone for scalar skills, periodic for \SkillRotation) may help extrapolation (Appendix~\ref{appendix:embedding_manifold} illustrates this for \SkillSize).

\paragraph{Conditioning mechanism.}
We test two methods for injecting condition information: direct concatenation of condition and image tokens, and AdaLN~\citep{peebles2023scalable}, which modulates network activations based on conditions.
Both succeed on training values, but neither extrapolates better on every skill.
AdaLN improves mean \SkillPosition\ and \SkillCount\ extrapolation ($61.5\%$ vs.\ $29.7\%$ and $72.5\%$ vs.\ $52.5\%$), while concatenation is stronger on \SkillSize\ ($51.5\%$ vs.\ $31.0\%$) and \SkillRotation\ ($68.0\%$ vs.\ $58.5\%$).
We use concatenation in the baseline because it has higher interpolation accuracy on average and is stronger on \SkillSize\ and \SkillRotation\ extrapolation, but AdaLN still performs competitively.

\paragraph{Pixel vs.\ latent space.}
We compare training the model directly in pixel space with training in the latent space of a pre-trained VAE~\citep{rombach2022high}.
Pixel-space training improves \SkillSize\ extrapolation ($51.5\%$ vs.\ $7.1\%$), \SkillPosition\ extrapolation ($29.7\%$ vs.\ $0.0\%$), and \SkillRotation\ extrapolation ($68.0\%$ vs.\ $21.2\%$).

Sequence length may account for part of the extrapolation gap.
The latent model sees 16 tokens per image, while the pixel model sees 1024.
With patch size 1, a latent model with 64 tokens reaches $36.1\%$ on \SkillSize\ extrapolation, up from $7.1\%$ with 16 tokens, but only $18.5\%$ on \SkillRotation\ (Appendix~\ref{appendix:token_matched}).
At $128\times128$, a latent model with 64 tokens is close to the pixel models on \SkillSize\ and \SkillPosition\ extrapolation (Appendix~\ref{appendix:hires}).
We keep pixel space as the default, although a latent model with enough tokens may perform similarly on \SkillSize.

\paragraph{Algorithm and architecture.}
We also change the generative algorithm to flow matching~\citep{lipman2023flow} and the backbone from DiT-S/2 to a U-Net~\citep{ronneberger2015u,ho2020denoising}.
For controlled comparison, we hold the data and optimization settings fixed.
The U-Net matches DiT-S/2 at about $22$M parameters (Appendix~\ref{appendix:architecture}).
Flow matching has lower mean extrapolation accuracy than the baseline on every skill, including \SkillRotation\ ($51.0\%$ vs.\ $68.0\%$).
The U-Net loses most of \SkillRotation\ extrapolation ($11.2\%$, the same as the nearest-training baseline) and is weaker on \SkillSize\ extrapolation ($21.2\%$) but is comparatively strong on \SkillCount\ extrapolation ($79.6\%$), so an architecture that helps one skill can hurt another.

\paragraph{Model size.}
We further explore whether generalization can be improved by scaling up model size.
Scaling the backbone from DiT-S/2 to DiT-L (${\sim}306$M parameters) improves mean \SkillCount\ and \SkillPosition\ extrapolation ($72.1\%$ vs.\ $52.5\%$ and $42.5\%$ vs.\ $29.7\%$), but does not extend extrapolation on \SkillSize\ ($44.5\%$ vs.\ $51.5\%$) or \SkillRotation\ ($62.0\%$ vs.\ $68.0\%$).
The capacity ablations are discussed in Appendix~\ref{appendix:dit_large_results}.

\section{Compositional Generalization}
\label{sec:compositional}

\begin{figure}[t]
\centering
\includegraphics[width=\textwidth]{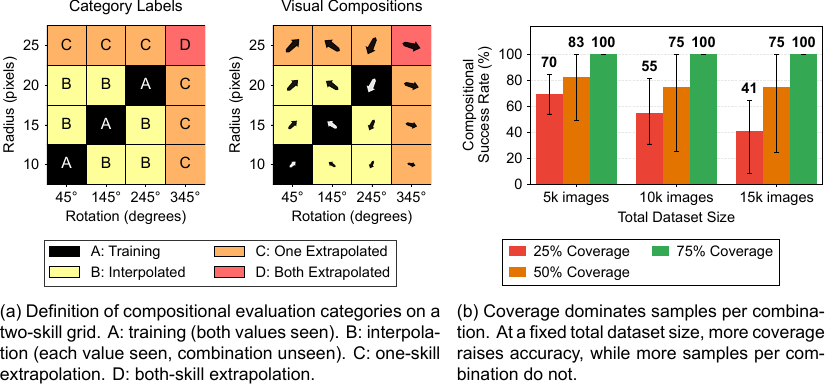}
\caption{\textbf{Covering more skill combinations helps compositional generalization more than adding samples per combination.} In (b), \SkillColor~+~\SkillShape\ models are trained on four shapes and eight colors, with 5k, 10k, or 15k images in total. Bars show the mean joint \CatB\ accuracy (both skills correct) over three seeds, and error bars show the range across seeds.}
\label{fig:compositional_combined}
\end{figure}

\begin{figure*}[t]
\centering
\includegraphics[width=0.92\textwidth]{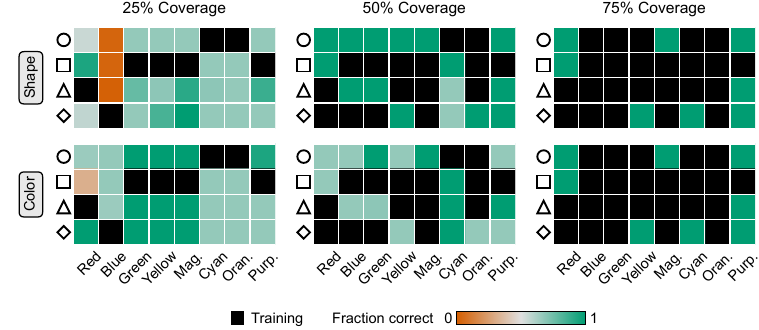}
\caption{\textbf{On \SkillColor~+~\SkillShape, held-out \SkillShape\ accuracy rises with coverage, and at 75\% coverage both skills are correct on every held-out combination.}
Results are for the 10k-image models of Figure~\ref{fig:compositional_combined}, with one cell per combination. The top row scores \SkillShape\ and the bottom row scores \SkillColor. Black cells are training combinations, and held-out cells are colored by the fraction of correct samples, averaged over the three seeds.}
\label{fig:compositional_heatmap}
\end{figure*}

\begin{findingbox}
\textbf{Finding 2.} For compositional generalization, the number of distinct skill combinations seen in training matters more than dataset size.
\end{findingbox}

Section~\ref{sec:single_property} studies one skill at a time, but controllable generation usually asks for several skills in the same image.
Combining skills in this way is closely related to the compositional generalization studied in controlled benchmarks for systematicity~\citep{johnson2017clevr,lake2018generalization,keysers2020measuring} and is central to skill-based evaluation of modern text-to-image models~\citep{huang2023t2i,ghosh2023geneval}.
We study compositional generalization by enumerating the full grid of skill values, training on only a fraction of the possible combinations, and evaluating whether models generate correct held-out combinations within range, as well as combinations that move beyond the joint training support.
Additional results can be found in Appendix~\ref{appendix:compositional}.

\paragraph{Setup.}
For each skill pair, we construct a discrete grid of combinations over a small set of values per skill (Appendix~\ref{appendix:datasets}).
We vary two aspects of the training distribution: (1) coverage, the fraction of unique combinations used in training, and (2) samples per combination (density), the number of images drawn for each training combination.
In the default setting (Table~\ref{tab:compositional_results}), every training combination receives the same number of images. In the coverage versus density experiment of Figures~\ref{fig:compositional_combined} and~\ref{fig:compositional_heatmap}, we instead hold the total dataset size fixed.
We also note that training combinations are sampled so that every skill value appears in at least one training combination (otherwise a held-out combination would test an unseen value rather than an unseen pairing).

We evaluate generated images with skill-specific metrics analogous to those used throughout the paper (Appendix~\ref{appendix:eval_protocol}) and group test queries into the categories defined in Figure~\ref{fig:compositional_combined}.
Category \CatA\ denotes the in-distribution baseline averaged across the pair's two skills (training values, training combinations).
Category \CatB\ evaluates compositional interpolation, where each skill value lies within its training range but the specific combination is unseen.
Category \CatC\ evaluates one-skill extrapolation, where one skill moves outside its training range while the other remains in range (we indicate which skill is out-of-distribution).
Category \CatD\ evaluates both-skill extrapolation, where both skills move outside their training ranges.

\paragraph{Coverage versus density.}
At a fixed image budget, seeing a wider variety of combinations helps composition more than seeing the same combinations more often.
Figure~\ref{fig:compositional_combined} reports joint \CatB\ accuracy (both \SkillColor\ and \SkillShape\ correct) at total budgets of 5k, 10k, and 15k images while varying coverage, so it isolates diversity of combinations from the number of samples per combination. We note these are separate models from those of Table~\ref{tab:compositional_results}, whose 25\% \SkillColor~+~\SkillShape\ split trains on 11 of the 32 combinations with about 3.4k images in total.
Moving toward higher coverage consistently raises \CatB\ accuracy, even though the number of samples per seen combination must fall to keep the total budget fixed.
By contrast, increasing the number of samples per combination at fixed coverage yields no consistent gain.
Each additional training combination shows the model another pairing of skill values, while extra samples of the same combination repeat a pairing the model has already seen.
Figure~\ref{fig:compositional_heatmap} shows this at the cell level for the 10k-image models: held-out \SkillShape\ accuracy grows with coverage, and both skills are correct on every held-out combination at 75\% coverage.

We also train \SkillColor~+~\SkillShape\ models with $S \in \{4, 8\}$ shapes and $C \in \{8, 16\}$ colors at three coverages, using about $15$k training images in every setting.
Each shape then appears in $k$ training combinations on average, where $k$ is the coverage times $C$.
Accuracy on held-out combinations is near $40\%$ when $k = 2$, between $59\%$ and $100\%$ when $k = 4$, and $100\%$ whenever $k \geq 6$ (Appendix~\ref{appendix:k_analysis}).
At a fixed image budget, more coverage and more colors therefore both raise held-out accuracy.

\begin{table*}[t]
\centering
\scriptsize
\setlength{\tabcolsep}{2.5pt}
\begin{tabular}{@{}lccccccccc@{}}
\toprule
\textbf{Visual Skills} & \textbf{Cov.} & \textbf{Cat.} & \textbf{\CatA\ (\%)} &
\textbf{\SkillShape\ (\%)} & \textbf{\SkillColor\ (\%)} & \textbf{\SkillPosition\ (\%)} & \textbf{\SkillSize\ (\%)} & \textbf{\SkillCount\ (\%)} & \textbf{\SkillRotation\ (\%)} \\
\midrule

\multirow{3}{*}{\SkillColor~+~\SkillShape} & 25\% & \CatB & 100.0 & $93.2_{\pm 14.9}$ & $94.5_{\pm 6.5}$ & \NA & \NA & \NA & \NA \\
 & 50\% & \CatB & 100.0 & $98.0_{\pm 3.1}$ & $95.6_{\pm 7.0}$ & \NA & \NA & \NA & \NA \\
 & 75\% & \CatB & 100.0 & $100.0_{\pm 0.0}$ & $100.0_{\pm 0.0}$ & \NA & \NA & \NA & \NA \\
\midrule

\multirow{3}{*}{\SkillColor~+~\SkillCount} & 25\% & \CatB & 100.0 & \NA & $\fail{79.9_{\pm 22.1}}$ & \NA & \NA & $\fail{38.2_{\pm 42.2}}$ & \NA \\
 & 50\% & \CatB & 100.0 & \NA & $96.1_{\pm 3.7}$ & \NA & \NA & $\fail{37.7_{\pm 3.5}}$ & \NA \\
 & 75\% & \CatB & 100.0 & \NA & $99.9_{\pm 0.3}$ & \NA & \NA & $97.4_{\pm 6.5}$ & \NA \\
\midrule

\multirow{3}{*}{\SkillShape~+~\SkillCount} & 25\% & \CatB & 100.0 & $\fail{13.3_{\pm 11.4}}$ & \NA & \NA & \NA & $\fail{69.6_{\pm 17.8}}$ & \NA \\
 & 50\% & \CatB & 100.0 & $\fail{13.7_{\pm 10.6}}$ & \NA & \NA & \NA & $\fail{83.7_{\pm 15.4}}$ & \NA \\
 & 75\% & \CatB & 100.0 & $\fail{10.0_{\pm 14.2}}$ & \NA & \NA & \NA & $\fail{46.4_{\pm 27.4}}$ & \NA \\
\midrule

\multirow{3}{*}{\SkillColor~+~\SkillPosition} & 25\% & \CatB & 100.0 & \NA & $97.4_{\pm 3.8}$ & $99.5_{\pm 1.2}$ & \NA & \NA & \NA \\
 & 50\% & \CatB & 100.0 & \NA & $100.0_{\pm 0.0}$ & $100.0_{\pm 0.0}$ & \NA & \NA & \NA \\
 & 75\% & \CatB & 100.0 & \NA & $100.0_{\pm 0.0}$ & $100.0_{\pm 0.0}$ & \NA & \NA & \NA \\
\midrule

\multirow{6}{*}{\SkillPosition~+~\SkillSize} & 25\% & \CatB & 99.3 & \NA & \NA & $99.8_{\pm 0.6}$ & $\fail{46.3_{\pm 8.8}}$ & \NA & \NA \\
 & 50\% & \CatB & 99.9 & \NA & \NA & $100.0_{\pm 0.0}$ & $96.0_{\pm 3.9}$ & \NA & \NA \\
 & 75\% & \CatB & 99.9 & \NA & \NA & $100.0_{\pm 0.0}$ & $99.7_{\pm 0.6}$ & \NA & \NA \\
 & 75\% & \CatC{\SkillSize} & 99.9 & \NA & \NA & $100.0_{\pm 0.0}$ & $\fail{21.8_{\pm 19.1}}$ & \NA & \NA \\
 & 75\% & \CatC{\SkillPosition} & 99.9 & \NA & \NA & $\fail{3.7_{\pm 9.0}}$ & $\fail{8.9_{\pm 10.0}}$ & \NA & \NA \\
 & 75\% & \CatD & 99.9 & \NA & \NA & $\fail{1.1_{\pm 2.7}}$ & $\fail{2.8_{\pm 6.7}}$ & \NA & \NA \\
\midrule

\multirow{3}{*}{\SkillPosition~+~\SkillShape} & 25\% & \CatB & 100.0 & $\fail{12.9_{\pm 8.0}}$ & \NA & $97.6_{\pm 2.5}$ & \NA & \NA & \NA \\
 & 50\% & \CatB & 100.0 & $\fail{54.6_{\pm 18.4}}$ & \NA & $99.6_{\pm 0.9}$ & \NA & \NA & \NA \\
 & 75\% & \CatB & 100.0 & $100.0_{\pm 0.0}$ & \NA & $100.0_{\pm 0.0}$ & \NA & \NA & \NA \\
\midrule

\multirow{3}{*}{\SkillSize~+~\SkillShape} & 25\% & \CatB & 100.0 & $\fail{34.4_{\pm 13.6}}$ & \NA & \NA & $\fail{26.2_{\pm 19.1}}$ & \NA & \NA \\
 & 50\% & \CatB & 100.0 & $\fail{82.5_{\pm 11.6}}$ & \NA & \NA & $\fail{50.5_{\pm 10.5}}$ & \NA & \NA \\
 & 75\% & \CatB & 100.0 & $93.7_{\pm 9.8}$ & \NA & \NA & $\fail{72.3_{\pm 12.9}}$ & \NA & \NA \\
\midrule

\multirow{6}{*}{\SkillPosition~+~\SkillRotation} & 25\% & \CatB & 100.0 & \NA & \NA & $\fail{64.9_{\pm 12.9}}$ & \NA & \NA & $\fail{5.6_{\pm 6.4}}$ \\
 & 50\% & \CatB & 100.0 & \NA & \NA & $95.3_{\pm 7.8}$ & \NA & \NA & $\fail{73.8_{\pm 28.8}}$ \\
 & 75\% & \CatB & 100.0 & \NA & \NA & $100.0_{\pm 0.0}$ & \NA & \NA & $100.0_{\pm 0.0}$ \\
 & 75\% & \CatC{\SkillPosition} & 100.0 & \NA & \NA & $\fail{0.8_{\pm 1.9}}$ & \NA & \NA & $\fail{54.4_{\pm 38.1}}$ \\
 & 75\% & \CatC{\SkillRotation} & 100.0 & \NA & \NA & $100.0_{\pm 0.0}$ & \NA & \NA & $\fail{0.0_{\pm 0.0}}$ \\
 & 75\% & \CatD & 100.0 & \NA & \NA & $\fail{2.2_{\pm 5.2}}$ & \NA & \NA & $\fail{1.5_{\pm 2.1}}$ \\

\bottomrule
\end{tabular}
\vspace{0.35em}
\caption{\textbf{Compositional interpolation improves with coverage for most two-skill pairs.}
Entries are per-skill accuracy (\%), the mean over six seeds with the standard deviation as a subscript. The \textbf{Cat.} column gives the test category from Figure~\ref{fig:compositional_combined}: \CatB\ is interpolation, \CatC\ moves the skill in parentheses out of range, and \CatD\ moves both skills out of range. The \CatA\ column is the in-distribution baseline averaged over the pair's two skills, and \fail{gray} marks accuracy below 90\%.}
\label{tab:compositional_results}
\end{table*}

\paragraph{Coverage across pairs.}
For most skill pairs, compositional interpolation improves with coverage (Table~\ref{tab:compositional_results}).
For example, \SkillCount\ accuracy on \SkillColor~+~\SkillCount\ rises from 38.2\% at 25\% coverage to 97.4\% at 75\%, and \SkillColor\ from 79.9\% to 99.9\%.
Continuous pairs also improve with coverage, e.g., \SkillSize\ accuracy on \SkillPosition~+~\SkillSize\ rises from 46.3\% at 25\% coverage to 99.7\% at 75\%.
\SkillShape~+~\SkillCount\ is the only pair where neither skill improves with coverage: at 75\% coverage, it reaches 10.0\% \SkillShape\ and 46.4\% \SkillCount\ accuracy, and most of its samples draw touching objects that merge into one component (Appendix~\ref{appendix:compositional}).
We further analyze the failure mode for \SkillSize~+~\SkillShape\ in Appendix~\ref{appendix:size_shape_failure}.
We discuss the out-of-distribution splits of \SkillPosition~+~\SkillSize\ and \SkillPosition~+~\SkillRotation\ in Section~\ref{sec:distribution_learning}.

\section{One Skill Out of Range}
\label{sec:distribution_learning}

\begin{findingbox}
\textbf{Finding 3.} Basic visual skills mostly fail independently, but an out-of-range \SkillPosition\ also lowers accuracy on some of the skills that stay in range.
\end{findingbox}

Compositional interpolation tests unseen combinations of in-range skill values.
We now move one skill outside its training range and test whether the other skills stay correct.
A model that learns each visual skill independently would degrade primarily on the out-of-distribution skill while preserving the accuracy of the remaining in-range skills.
Diffusion models condition on all skills jointly during denoising, so the resulting representations can become coupled.

\paragraph{Two skills.}
With two skills, the coupling runs mainly from \SkillPosition\ to the other skill (\Cref{tab:compositional_results}, 75\% coverage).
When only \SkillPosition\ is OOD, accuracy on the in-range skill drops from near $100\%$ to $8.9\%$ for \SkillSize\ and $54.4\%$ for \SkillRotation, although the requested sizes and angles are within their training ranges.
An OOD \SkillPosition\ also hurts \SkillSize\ when \SkillSize\ is itself OOD, since for the same queried sizes its accuracy falls from $21.8\%$ to $2.8\%$ once \SkillPosition\ also leaves the training range (\CatD).
In the opposite direction, \SkillPosition\ is unaffected, at $100.0\%$ when \SkillSize\ is OOD and $100.0\%$ when \SkillRotation\ is OOD. In the other two-skill models, an OOD \SkillSize\ or \SkillCount\ does not lower \SkillShape\ or \SkillColor\ either (Appendix~\ref{appendix:multi_prop}).

\begin{table}[t]
\centering
\small
\begin{tabular}{@{}lcccc@{}}
\toprule
\textbf{Split} & \textbf{\SkillSize} & \textbf{\SkillPosition} & \textbf{\SkillShape} & \textbf{\SkillColor} \\
\midrule
In range (\CatB) & 100.0 & 100.0 & 100.0 & 100.0 \\
\SkillSize\ OOD (\CatC) & \fail{54.6} & 100.0 & 100.0 & 100.0 \\
\SkillPosition\ OOD (\CatC) & \fail{79.5} & \fail{0.0} & \fail{32.0} & 100.0 \\
\bottomrule
\end{tabular}
\caption{\textbf{Skill interdependence in a four-skill model.}
Accuracy (\%) of each skill for one model trained on \SkillSize, \SkillPosition, \SkillShape, and \SkillColor\ together at $75\%$ coverage, unlike the two-skill models of Table~\ref{tab:compositional_results}. \fail{Gray} marks accuracy below $90\%$. The two three-skill runs show the same pattern (Appendix~\ref{appendix:multi_prop}).}
\label{tab:interdependence_main}
\end{table}

\paragraph{Three and four skills.}
Models trained on \SkillSize, \SkillPosition, and \SkillShape\ and on these three plus \SkillColor\ show the same pattern (\Cref{tab:interdependence_main}; Appendix~\ref{appendix:multi_prop}).
An OOD \SkillPosition\ lowers \SkillShape\ accuracy from $100\%$ on in-range combinations to $20.4\%$ and $59.9\%$ in the two three-skill runs and to $32.0\%$ in the four-skill model, although the requested shapes are in range.
An OOD \SkillSize\ leaves \SkillPosition\ and \SkillShape\ at $100\%$.
In the four-skill model, \SkillColor\ stays at $100\%$ on every split, so the coupling holds for some skills but not all.
\section{Related Work}
\label{sec:related_work}

\paragraph{Evaluating Visual Generative Models.}
Generative models are commonly evaluated with distribution-level metrics such as IS and FID~\citep{salimans2016improved,heusel2017gans,sajjadi2018assessing,kynkaanniemi2019improved,naeem2020reliable}, and text-to-image models with human preference~\citep{wu2023human,jiang2024genai} and benchmarks for prompt following, attribute binding, and counting~\citep{lee2023holistic,bakr2023hrs,hu2023tifa,huang2023t2i,ghosh2023geneval,hu2024ella,chang2025oneig,guo2025text,wu2024conceptmix}.
These evaluations operate on pre-trained models whose training distributions are not controlled, and each score inherits the errors of its detector or VQA model.
Studies of compositional generalization instead use synthetic data with known factors, in language and visual reasoning~\citep{lake2018generalization,keysers2020measuring,johnson2017clevr} and in representation learning~\citep{matthey2017dsprites,bordes2023pug,uselis2025data}.
VGL also trains on carefully controlled data, so every test value is known to lie inside or outside the training range, and scores each skill with task-specific, rule-based metrics.

\paragraph{Memorization and Generalization.}
Modern models can generalize well while still memorizing their training data~\citep{zhang2017understanding,arpit2017closer}, and in diffusion models memorization depends on training-set size, data duplication, text conditioning, and classifier-free guidance~\citep{carlini2023extracting,somepalli2023diffusion,somepalli2023understanding,kadkhodaie2024generalization,jain2025classifier}.
Other work trains diffusion models on synthetic data to study how compositional abilities emerge~\citep{okawa2023compositional,park2024emergence} and when a model can generate more objects than it saw in training~\citep{bradley2025local}.
VGL complements these efforts with a controlled setting for measuring when diffusion models extrapolate a skill versus when they fall back to nearest-support solutions.

\section{Discussion and Conclusion}
\label{sec:conclusion}

We introduced VGL to explore whether diffusion models learn basic visual skills or mainly reproduce what their training data contain.
In our controlled setting, we find models mostly handle unseen values and combinations within what they were trained on, but rarely carry a skill far beyond it.
If larger models behave the same way, data providers and model trainers may need to curate data that cover the skill values and combinations users ask for, rather than rely on more data alone, and evaluations should test outside that coverage. We hope VGL can serve as a starting point for studying these questions in richer settings.

\paragraph{Limitations and future work.}
VGL relies on controlled synthetic data, and we mainly evaluate a DiT backbone, so how far these findings carry over to natural images remains open.
As a first step, we repeat the single-skill experiments on textured renders (Appendix~\ref{appendix:realistic}) and at higher resolution (Appendix~\ref{appendix:hires}), and probe eight pre-trained text-to-image models with rule-based extractors (Appendix~\ref{appendix:t2i}).
Going forward, we aim to scale VGL to richer scenes, broaden architectures beyond DiT, and apply the same skill-centric probes to larger text-to-image models, to see whether the coverage and interdependence patterns we observe hold at larger scales and in more realistic settings.

\begin{ack}
This material is based upon work supported by the U.S. National Science Foundation Graduate Research Fellowship Program under Grant No.~DGE-2630152. Any opinions, findings, and conclusions or recommendations expressed in this material are those of the author(s) and do not necessarily reflect the views of the National Science Foundation. This work received no other third-party funding. The authors declare no competing interests.
\end{ack}

\bibliographystyle{plainnat}
\bibliography{main}

\clearpage
\appendix
\crefalias{section}{appendix}
\crefalias{subsection}{subappendix}
\renewcommand*\contentsname{\Large Appendix}
{
  \hypersetup{linkcolor=black}
  \tableofcontents
}

\addtocontents{toc}{\protect\setcounter{tocdepth}{2}}
\clearpage
\section{Experimental Details}
\label{appendix:experimental_details}

\subsection{Dataset specifications}
\label{appendix:datasets}

We define six visual skills and their value ranges.
Table~\ref{tab:skill_domains} summarizes the complete domain specification for each skill.

\begin{table}[htbp]
\centering
\small
\setlength{\tabcolsep}{3.5pt}
\renewcommand{\arraystretch}{1.05}
\begin{tabularx}{\linewidth}{@{}l l >{\raggedright\arraybackslash}X >{\raggedright\arraybackslash}X@{}}
\toprule
\textbf{Visual Skill} & \textbf{Type} & \textbf{Range / Values} & \textbf{Value grid} \\
\midrule
\SkillSize & Continuous & $[5.0, 20.0]$ pixels & 16 values (step 1.0) \\
\SkillPosition & Continuous (2D) & $[-18, 18] \times [-18, 18]$ & $32 \times 32$ grid (1024 positions) \\
\SkillRotation & Continuous & $[45^\circ, 315^\circ]$ & 271 values (step 1$^\circ$) \\
\SkillCount & Discrete & $\{2, 3, \ldots, 7\}$ objects & 6 values \\
\SkillShape & Categorical & circle, square, triangle, diamond & 4 classes \\
\SkillColor & Categorical & red, blue, green, yellow, magenta, cyan, orange, purple & 8 classes \\
\bottomrule
\end{tabularx}
\vspace{0.35em}
\caption{\textbf{Visual skill domains and value grids.}
The value grid column gives the number of values in each domain, and positions form a uniform grid over the listed range.}
\label{tab:skill_domains}
\end{table}

\paragraph{Rendering pipeline.}
All images are rendered programmatically using Python with PIL (Pillow) and NumPy at $64 \times 64$ pixels, except the $128 \times 128$ runs of Appendix~\ref{appendix:hires}.
Each image is initially rendered at $256 \times 256$ pixels and downsampled to $64 \times 64$ using Lanczos resampling; the single-skill \SkillCount\ images are instead drawn directly at $64 \times 64$.
The eight \SkillColor\ classes use standard RGB values.
These are red (255,0,0), blue (0,0,255), green (0,255,0), yellow (255,255,0), magenta (255,0,255), cyan (0,255,255), orange (255,128,0), and purple (128,0,255).
Background colors are sampled from whites, light grays, and pastels with RGB values typically above 200 per channel.

Shapes are rendered as follows.
Circles use PIL's ellipse primitive with diameter $= 2 \times$ size.
Squares are axis-aligned with side length $= 2 \times$ size.
Triangles are equilateral with side length $= 2 \times$ size.
Diamonds are squares rotated by $45^\circ$ with their vertices at distance size from the center.
All shapes use solid fills with no borders.
The only object drawn at varying angles is the arrow used in the single-skill \SkillRotation\ experiments, which the \SkillPosition~+~\SkillRotation\ pair also uses.

In the single-skill datasets of Table~\ref{tab:single_property_results}, each object takes a random color from a separate muted eight-color palette, perturbed by Gaussian noise with $\sigma = 10$ per RGB channel (clipped to $[0, 255]$).
Gaussian noise with $\sigma = 5$ is then added to the entire image.
The \SkillRotation\ dataset is the exception, with the arrow drawn in (255,100,100) on white and no perturbation.

The compositional datasets of Table~\ref{tab:compositional_results} and Appendix~\ref{appendix:multi_prop} draw objects in red unless \SkillColor\ is one of the skills, add uniform jitter of up to $\pm 15$ to each object color channel, and add Gaussian image noise with $\sigma \approx 5$.
The \SkillColor~+~\SkillShape\ datasets of Figure~\ref{fig:compositional_combined} instead shift each color channel of the whole image by a uniform offset in $[-8, 8]$ and add uniform pixel noise in $[-3, 3]$.
All color and noise perturbations are applied during dataset generation only; the single-skill \SkillSize\ and \SkillCount\ models also flip training images horizontally at random.

\paragraph{Train and test splits.}
For single-skill experiments, we discretize the skill's continuous range and treat each discrete value as a class.
Training sets use every value of the grid in Table~\ref{tab:skill_domains}; the rotation ablations of \Cref{appendix:rotation_ablations} instead vary the range and step.
Interpolation samples are midpoints between consecutive training values.
Extrapolation out-of-distribution samples are values outside the training range but within the valid physical range.
For \SkillSize, the extrapolation range stops at radius 30, where the circle comes within 2~px of each edge of the $64 \times 64$ canvas.
For \SkillPosition, Table~\ref{tab:single_property_results} continues the training lattice outward in its 1.16~px steps and scores the 320 points that lie at least 4~px outside the training square in $L_\infty$ distance (4.6 or 5.8~px).

For compositional experiments involving multiple skills, coverage is defined over the Cartesian product of skill values.
In the two-skill grids of Table~\ref{tab:compositional_results}, \SkillSize\ uses radii $\{6, 8, \ldots, 18\}$ px, trained on $\{6, 8, \ldots, 14\}$ with $\{16, 18\}$ held out.
\SkillPosition\ uses $\{-14, -7, 0, 7, 14\}^2$ px, trained on $x \leq 7$ with the $x = 14$ column held out.
\SkillRotation\ is trained on $\{45^\circ, 135^\circ, 225^\circ, 315^\circ\}$ with $\{0^\circ, 15^\circ, 30^\circ, 330^\circ, 345^\circ\}$ held out, and \SkillCount\ uses $\{1, 2, 3, 4\}$, trained on $\{1, 2, 3\}$ with 4 held out.
\SkillShape\ and \SkillColor\ use all their classes in training.
A skill that a pair does not vary is fixed at radius 10 px, the image center, a circle (the arrow for \SkillPosition~+~\SkillRotation), red, one object, and $0^\circ$.
In \SkillColor~+~\SkillCount\ and \SkillShape~+~\SkillCount, the objects have radius 5 px and sit in one horizontal row through the image center, with centers 12.5 px apart.

Using seed 42, we randomly sample a fraction $p\%$ of all combinations of training values without replacement.
We then add one combination for each skill value that is still missing, so that every value appears in training.
Each added combination pairs the missing value with the first training value of the other skill (e.g., radius 6, one object, $45^\circ$, a circle, or red).
In the datasets of Table~\ref{tab:compositional_results}, this raises the realized coverage above its nominal value, most for the smallest grid, \SkillShape~+~\SkillCount, which reaches 50\% at a nominal 25\% and 58\% at a nominal 50\%. At a nominal 25\%, \SkillColor~+~\SkillShape\ trains on 11 of its 32 combinations.
The test splits follow Categories \CatA\ to \CatD\ of Section~\ref{sec:compositional}.

For the single-skill experiments of Table~\ref{tab:single_property_results}, 10{,}000 training images, generated with seed 42, are split evenly across the training values (625 per \SkillSize\ value, about 10 per \SkillPosition\ grid point, about 37 per \SkillRotation\ angle, and about 1{,}667 per \SkillCount\ value).
For the two-skill results in Table~\ref{tab:compositional_results}, each training combination receives the same number of samples, from 50 to 357 depending on the pair (312 for \SkillColor~+~\SkillShape).
The total therefore grows with coverage, and so does the number of optimizer steps, since training runs for a fixed 1000 epochs.

The fixed-budget comparisons in Figure~\ref{fig:compositional_combined} and Table~\ref{tab:k_analysis} instead fix the total training size (for example 15{,}000 images) and distribute samples uniformly across selected combinations.
In Figure~\ref{fig:compositional_combined}(b), the three coverages train on 8, 16, and 24 of the 32 \SkillColor~+~\SkillShape\ combinations, with 208 to 1{,}875 images per training combination.

\subsection{Count object layout and training schedule}
\label{appendix:count_geometry}

The \SkillCount\ dataset of Table~\ref{tab:single_property_results} places radius-4 circles, drawn 9 px wide, at distinct sites of a 16-px lattice, which leaves 7 px of clearance between neighbors.
\SkillCount\ is the only skill still improving at 1000 epochs: on this layout the DiT-S/2 baseline reaches $69.7_{\pm 11.9}\%$ training accuracy at 1000 epochs (78k steps) and $100.0_{\pm 0.0}\%$ at 3000 (234k steps), with three seeds each.
For the model shown in Figure~\ref{fig:qualitative}, training accuracy is $14\%$ at 30k steps, $95\%$ at 120k, and $100\%$ at about 200k steps, after which it is flat, while the training loss stays at ${\approx}0.017$ over the same interval.
We use accuracy rather than loss as the convergence criterion for this skill and train it for 3000 epochs.

\subsection{Model architecture and training}
\label{appendix:architecture}

We use DiT-S/2 (Diffusion Transformer, Small variant) with 12 layers, hidden dimension 384, 6 attention heads, and patch size 2, yielding $32 \times 32$ patches for $64 \times 64$ images.
The model has approximately 22M parameters without the VAE.
The U-Net variants in Table~\ref{tab:single_property_results} have $22.35$M parameters for \SkillSize\ and \SkillCount, $22.39$M for \SkillPosition, and $22.96$M for \SkillRotation, against $22.20$M for DiT-S/2 ($22.44$M for \SkillPosition).
The AdaLN variants have $33.12$M parameters ($33.36$M for \SkillPosition), because each block adds a modulation layer.

Each visual skill is embedded into the model's hidden dimension (384).
Continuous skills use a linear embedding $\mathbf{e} = \mathrm{MLP}(\mathbf{W}v + \mathbf{b})$, where a two-layer MLP maps the 256-dimensional projection to the hidden size.
For \SkillSize\ and \SkillCount, $v$ is the scalar value and $\mathbf{W} \in \mathbb{R}^{256 \times 1}$; \SkillPosition\ embeds $x$ and $y$ separately through two such projections and prepends both tokens.
\SkillRotation\ uses $\mathbf{W} \in \mathbb{R}^{256 \times 2}$ on $[\sin\theta, \cos\theta]$, so its input is periodic by construction.
The compositional models instead embed $(x, y)$ jointly as one token with $\mathbf{W} \in \mathbb{R}^{256 \times 2}$ and replace $\mathbf{W}v$ for \SkillRotation\ with a fixed 256-dimensional sinusoidal embedding of $\theta$ in radians, which is not periodic.
We use no periodic construction in these models so that the representation stays consistent across all skills.

The sinusoidal variant of Table~\ref{tab:single_property_results} replaces $\mathbf{W}v$ with $[\cos(\omega_k v), \sin(\omega_k v)]_{k<128}$, and the rotary variant replaces it with a learned unit vector whose coordinate pairs are rotated by the angles $\omega_k v$, where $\omega_k = 10000^{-k/128}$.
For \SkillRotation, $v$ is $\theta/360^\circ$ in the sinusoidal and $\theta$ in radians in the rotary embedding, with $\theta$ taken modulo $360^\circ$.
Shape and color use learned embedding tables where each category $c$ has a learnable vector $\mathbf{e}_c \in \mathbb{R}^{384}$.

We use concatenation conditioning (in-context conditioning~\citep{peebles2023scalable,bao2023all}), where skill embeddings are prepended as separate tokens to the image patch sequence.
The input sequence is $[\mathbf{e}_{\text{time}}, \mathbf{e}_{\text{skill}_1}, \mathbf{e}_{\text{skill}_2}, \ldots, \mathbf{x}_{\text{patch}_1}, \mathbf{x}_{\text{patch}_2}, \ldots]$.
The timestep embedding $\mathbf{e}_{\text{time}}$ (sinusoidal, 256-dimensional, projected to 384 dimensions) is always included.
Each active skill adds one token (position adds separate $x$ and $y$ tokens in the single-skill models).
Standard transformer self-attention processes all tokens jointly.
After the transformer, we discard conditioning tokens and apply a final linear projection to predict noise.
The U-Net has no token sequence, so it instead adds a learned projection of the skill embedding to the features of every block.

\paragraph{Training objective.}
We use standard DDPM training with a linear noise schedule over $T = 1000$ diffusion steps with $\beta_t$ from $\beta_1 = 0.0001$ to $\beta_T = 0.02$.
The model is trained to predict the noise $\mathbf{\epsilon}$ with a mean squared error loss,
$\mathcal{L} = \mathbb{E}_{\mathbf{x}_0, \mathbf{\epsilon}, t} [ \| \mathbf{\epsilon} - \mathbf{\epsilon}_\theta(\mathbf{x}_t, t, \text{cond}) \|^2 ]$.
It also predicts the variance (a 6-channel output for 3-channel RGB images, split into noise and log-variance predictions).
The loss adds the variational lower bound term with weight 0.001~\citep{nichol2021improved}.
The flow-matching models instead regress the velocity $\mathbf{x}_0 - \mathbf{\epsilon}$ along $\mathbf{x}_t = (1-t)\,\mathbf{\epsilon} + t\,\mathbf{x}_0$ with $t \sim \mathcal{U}[0, 1]$, which enters the DDPM timestep embedding without rescaling.

\paragraph{Optimization.}
We use the AdamW optimizer with $\beta_1 = 0.9$, $\beta_2 = 0.999$, $\epsilon = 10^{-8}$, constant learning rate $1 \times 10^{-4}$, and weight decay 0.0. The single-skill \SkillRotation\ models and the \SkillCount\ models of Table~\ref{tab:single_property_results} also clip the global gradient norm at 1.0.
Global batch size is 128.
The two \SkillSize\ models of Appendix~\ref{appendix:embedding_manifold}, the single runs of \Cref{tab:rotation_ablations}, and the compositional models other than the vocabulary sweep of Table~\ref{tab:k_analysis} use 64.
We train for 1000 epochs for all experiments except the \SkillCount\ models of Table~\ref{tab:single_property_results}, which use 3000 (Appendix~\ref{appendix:count_geometry}), the data-scaling runs, which use 78k to 80k steps (Appendix~\ref{appendix:data_scaling}), and the vocabulary sweep of Table~\ref{tab:k_analysis}, which uses 500.

The DiT-S/2 models of Tables~\ref{tab:single_property_results} and~\ref{tab:compositional_results} train on one NVIDIA L40 GPU, except the 3000-epoch \SkillCount\ models, which train on four L40 or RTX PRO 6000 GPUs; the DiT-L and U-Net models train on four to eight.
Runs with a smaller per-GPU micro-batch accumulate gradients so that the global batch size is unchanged.
We maintain an EMA copy of model weights with decay rate 0.9999 updated every training step.
Evaluation uses the EMA model, except for the two-skill models of Table~\ref{tab:compositional_results}, which are evaluated with the online (non-EMA) weights.
We do this because the EMA starts from the random initialization, which keeps a weight of $0.9999^{n}$ after $n$ steps ($0.12$ at 21k), and many of these runs train for only 21k to 45k steps. Thus, on the shortest runs, the EMA weights lower accuracy on the training combinations from $100\%$ to below $16\%$.

\subsection{Evaluation protocol}
\label{appendix:eval_protocol}

\paragraph{Metrics per skill.}
For \SkillSize, we threshold the grayscale image with Otsu's method and take the class on the frame border as background.
When Otsu's two classes are less than 20 gray levels apart, we instead threshold halfway between the median of the frame border and the gray level farthest from it.
This happens when Otsu's method splits the background noise around a very small disk.
We apply a $3\times3$ morphological closing and, to masks of at least 20 pixels, a $3\times3$ opening.
We then estimate size as $r_{\text{gen}} = \sqrt{A/\pi}$ from mask area $A$, render a ground-truth circle with expected size $r_{\text{true}}$, and compute IoU between generated and ground-truth masks (success criterion IoU at least 0.90, which for concentric disks allows a radius error of about $5\%$).
We report accuracy and mean IoU.

For \SkillPosition, we compute the binary mask centroid $(c_x, c_y)$ as center of mass, convert to image relative coordinates (center is (0,0), range $[-32,32]$), and compute Euclidean distance $d$ in pixels (success criterion $d \leq 2.0$ pixels).
We report accuracy and mean absolute error (MAE) in pixels. For the DiT-L and U-Net rows of Table~\ref{tab:single_property_results}, \SkillPosition\ training and interpolation accuracy is measured on one fixed random sample of 100 conditions per split, and extrapolation on all 320 conditions.

For \SkillRotation, we use template matching against arrow renders at whole-degree angles, searched coarse to fine in $10^\circ$, $3^\circ$, and $1^\circ$ steps, and compute circular angular error.
Because the templates are at whole degrees, interpolation midpoints are read $0.5^\circ$ off, so interpolation MAE cannot fall below $0.5^\circ$.
The success threshold is $\Delta\theta \leq 5^\circ$.
We report accuracy and MAE in degrees.

For \SkillCount, we blur the grayscale image and split it with Otsu's method; if the two classes are less than 20 gray levels apart, the frame is treated as blank and the count is $0$.
Otherwise the class on the frame border is background, and the object mask is thresholded halfway between Otsu's threshold and the background mean, which keeps light objects whole.
We separate touching objects by watershed segmentation~\citep{vincent1991watersheds} of the distance transform, with markers where the distance exceeds 0.6 of the largest distance in the same blob, count regions with area larger than 30 pixels, and require an exact match with no partial credit.
The area threshold is below the area of one object ($61$ px$^2$ as drawn at radius 4), so it removes speckle and keeps every object.
We report accuracy.

For \SkillShape, we classify the object by template matching against renders of each shape.
For \SkillColor, we compute mean RGB within the object mask and assign the closest color from the eight-color palette of Table~\ref{tab:skill_domains} using L2 distance.
The vocabulary sweep of Table~\ref{tab:k_analysis} averages over the mask eroded by one pixel and assigns the closest of the cell's $C$ colors.
We report accuracy for both shape and color.

\paragraph{Multi-skill metrics.}
For the two-, three-, and four-skill models (Tables~\ref{tab:compositional_results} and~\ref{tab:multi_prop_interdependence}, Figures~\ref{fig:compositional_combined}(b) and~\ref{fig:compositional_heatmap}), every generated sample is scored on its own against the values it was conditioned on, and accuracy is the fraction of samples that pass.
The object mask is the Otsu threshold of each pixel's RGB distance to the background color.
A sample in which at most 10 pixels differ from the background by more than 60 contains no object and fails every skill.

Without \SkillCount, the object is the largest connected component.
With \SkillCount, every connected component of at least 30 pixels is an object, and \SkillShape\ passes only when each has the requested shape, so touching objects that merge into one component fail \SkillShape.
In the \SkillCount\ pairs, \SkillColor\ is the mean over the whole object mask.
\SkillSize\ passes when the object's area is between $0.9$ and $1/0.9$ times the area of a render of the requested shape and size, which allows a size error of about $5\%$ (two concentric renders with this area ratio have an IoU of 0.90).

\SkillPosition\ passes when the object's centroid is within 2 pixels of the centroid of a render at the requested position, and \SkillRotation\ passes when template matching against a $1^\circ$ bank of arrow renders is within $5^\circ$.
\SkillCount\ counts connected components of at least 30 pixels and splits touching objects where the distance transform narrows to at most 0.75 of the smaller object's peak.
\SkillShape\ takes the best fit among circle, square, and triangle templates searched over orientation.
A square within $15^\circ$ of axis-aligned is a square and one within $15^\circ$ of $45^\circ$ is a diamond, a triangle must be within $15^\circ$ of upright, and elongated or poorly fitting objects match no shape.

\paragraph{Ground-truth check.}
On ground-truth renders, every metric above scores $100\%$ on each split of Table~\ref{tab:single_property_results}, of the multi-skill results, and of the vocabulary sweep of Table~\ref{tab:k_analysis}.
On the other \SkillSize\ and \SkillCount\ data of Appendix~\ref{appendix:additional}, including the textured renders, the pass rate is at least $99\%$ per split.

\paragraph{Nearest-training baseline.}
For each out-of-range \SkillSize\ or \SkillRotation\ query, the baseline returns the nearest training value, which is the closer end of the training range.
We score it in closed form, by the IoU of concentric circles with the queried and returned radii for \SkillSize\ and by the circular angular error for \SkillRotation; Figure~\ref{fig:radius_rotation} shows this baseline as the dashed line.
It scores $7.1\%$ on \SkillSize\ and $11.2\%$ on \SkillRotation\ extrapolation.
Applied to \SkillPosition\ and \SkillCount, the same rule scores $0\%$ on extrapolation, since no out-of-range query is within the success threshold of a training value.
The DiT-S/2 (baseline) row of Table~\ref{tab:single_property_results} is above it on extrapolation for all four skills.
On interpolation, the rule scores about $70\%$ for \SkillSize\ and $100\%$ for \SkillPosition\ and \SkillRotation, whose midpoints all lie within the success threshold of a training value.

\paragraph{Sampling procedure.}
At evaluation, we use DDPM ancestral sampling with 250 steps respaced from the 1000 training steps~\citep{nichol2021improved}; the flow-matching models instead integrate the learned velocity field with 250 Euler steps that evaluate the velocity at each step's midpoint time.
We initialize $\mathbf{x}_T \sim \mathcal{N}(0, \mathbf{I})$ and iteratively denoise.
For each test condition, we generate $N = 20$ samples and report metrics averaged over these samples.
For the single-skill DDPM models, the initial noise $\mathbf{x}_T$ uses a fixed seed per sample index.

\subsection{Classifier-free guidance ablation (\SkillSize)}
\label{appendix:cfg_size_ablation}

\begin{figure}[htbp]
\centering
\includegraphics[width=0.50\linewidth]{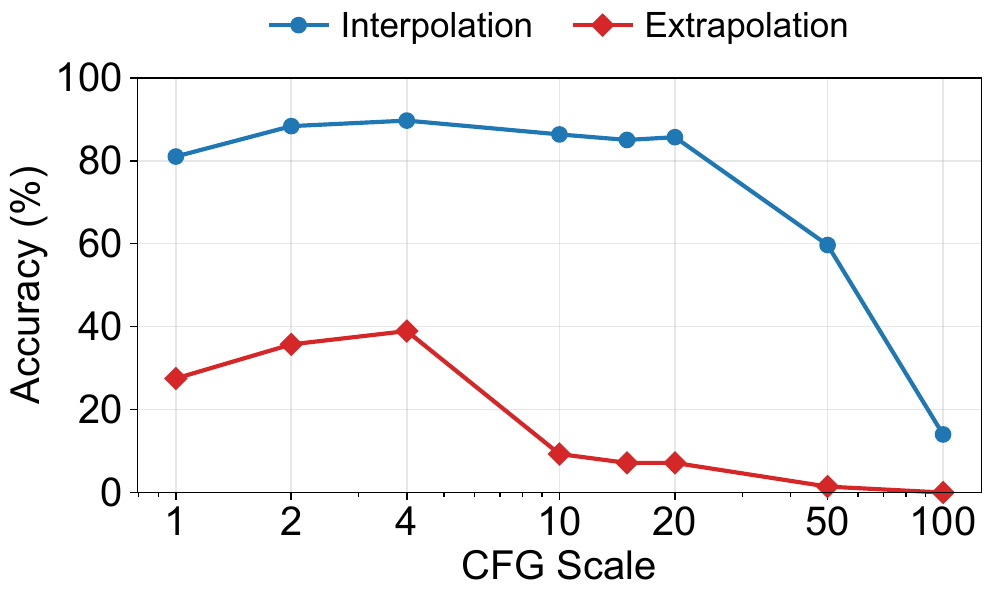}
\caption{\textbf{Classifier-free guidance does not remove the \SkillSize\ extrapolation limit.}
Accuracy of the \SkillSize\ baseline trained with conditioning dropout 0.1 (one seed), across inference guidance scales.
Scales up to $4$ raise accuracy on both splits but move the upper extrapolation limit only from radius 22 to 24, and larger scales lower both.}
\label{fig:cfg_scale_size_ablation}
\end{figure}

For the \SkillSize\ experiment, we also train the baseline of Table~\ref{tab:single_property_results} for classifier-free guidance (CFG), with conditioning dropout 0.1 on one seed, and evaluate how guidance strength affects accuracy on the same interpolation and extrapolation radii, with 20 samples per condition.
At inference time, we vary the guidance scale from 1 to 100 (\Cref{fig:cfg_scale_size_ablation}).
Scales up to $4$ improve both splits but extend \SkillSize\ extrapolation by only two radii, and from guidance scale 10 on, extrapolation accuracy stays below $10\%$; thus, guidance does not remove the extrapolation limit.
At every scale, this model extrapolates less accurately than the baseline trained without dropout ($51.5\%$).

\clearpage
\section{Rotation Ablations}
\label{appendix:rotation_ablations}

\paragraph{Training-coverage ablations.}
In these ablations, accurate rotation extrapolation needs a dense grid of training angles.
We evaluate six variants of DiT-S/2 trained on other subsets of the rotation skill space (\Cref{tab:rotation_ablations}).
Reducing angular coverage from $270^\circ$ to $180^\circ$ ($90^\circ$ to $269^\circ$ at a $1^\circ$ step) keeps the training angles accurate (MAE $0.17^\circ$), with $45.9\%$ extrapolation accuracy and $44.5^\circ$ MAE.
Its extrapolation angles reach $90^\circ$ beyond the training range, while those of the main configuration reach $45^\circ$.
Within $45^\circ$ of the training range, its extrapolation accuracy is $73.3\%$, inside the spread of the ten seeds of the main configuration ($32.3\%$ to $100.0\%$, mean $68.0\%$).
At a $1^\circ$ step, the run on $30^\circ$ to $329^\circ$ reaches $100.0\%$ extrapolation accuracy on angles up to $30^\circ$ from its training range; at the same distances, the ten seeds of the main configuration average $79.7\%$ ($47.9\%$ to $100.0\%$).

Value-grid spacing impacts generalization.
Models trained with $1^\circ$ increments substantially outperform those with $30^\circ$ increments, which fit their training angles but extrapolate poorly.
For the range starting at $45^\circ$, increasing the increment from $1^\circ$ to $30^\circ$ lowers extrapolation accuracy from $68.0\%$ (mean of the ten seeds, $32.3\%$ to $100.0\%$) to $14.7\%$.
The $30^\circ$ grid ends at $285^\circ$, but even within $10^\circ$ of its training range its accuracy is $70.5\%$, against $99.2\%$ on average for the ten seeds of the $1^\circ$ grid ($94.3\%$ to $100.0\%$).
We use the 10-seed analysis below to characterize seed sensitivity for our main configuration.

\begin{table*}[htbp]
\centering
\small
\renewcommand{\arraystretch}{1.05}
\begin{tabular}{@{}lcccccc@{}}
\toprule
 & & & \multicolumn{3}{c}{\textbf{MAE ($^\circ$)}} & \textbf{Acc.\ (\%)} \\
\cmidrule(lr){4-6} \cmidrule(lr){7-7}
\textbf{Training angles} & \textbf{Step} & $\boldsymbol{N}$ & \textbf{Train} & \textbf{Interp.} & \textbf{Extrap.} & \textbf{Extrap.} \\
\midrule
$30^\circ$ to $329^\circ$ & $1^\circ$ & 300 & 0.30 & 0.53 & 0.87 & 100.0 \\
$30^\circ$ to $300^\circ$ & $30^\circ$ & 10 & 0.10 & 10.6 & 26.4 & 11.2 \\
$45^\circ$ to $285^\circ$ & $30^\circ$ & 9 & 0.00 & 2.6 & 47.9 & 14.7 \\
$90^\circ$ to $269^\circ$ & $1^\circ$ & 180 & 0.17 & 0.51 & 44.5 & 45.9 \\
$90^\circ$ to $255^\circ$ & $15^\circ$ & 12 & 0.08 & 2.4 & 98.4 & 22.8 \\
$90^\circ$ to $240^\circ$ & $30^\circ$ & 6 & 0.17 & 3.1 & 82.2 & 8.2 \\
\midrule
$45^\circ$ to $315^\circ$ (main, 10 seeds) & $1^\circ$ & 271 & $0.30_{\pm 0.03}$ & $0.54_{\pm 0.01}$ & $19.2_{\pm 22.5}$ & $68.0_{\pm 28.9}$ \\
\bottomrule
\end{tabular}
\vspace{0.35em}
\caption{\textbf{Rotation training set ablations.} Each row is a DiT-S/2 model (linear embedding, concatenation conditioning, no conditioning dropout) trained on $N$ angles over the listed range and step. Columns give mean absolute error (MAE) on the training angles, their midpoints (Interp.), and every whole degree outside the training range (Extrap.), plus extrapolation accuracy at the $5^\circ$ threshold. Training accuracy is $100.0\%$ for every model. Rows above the line are one seed each, and the two at a $1^\circ$ step score Train and Interp.\ on 50 evenly spaced angles. The last row is the mean$\pm$std of the ten seeds of \Cref{tab:single_property_results}. For the same starting angle, a $1^\circ$ step gives lower interpolation and extrapolation errors than a coarser one.}
\label{tab:rotation_ablations}
\end{table*}

\paragraph{Per-seed behavior of the main configuration.}
For the main configuration ($45^\circ$ to $315^\circ$, $1^\circ$ step), we train ten seeds with identical data and schedule and report per-seed mean angular error and accuracy at the $5^\circ$ threshold in \Cref{tab:rotation_per_seed_main}.
Inside the training range, all ten seeds produce essentially identical numbers (training MAE $0.30 \pm 0.03^\circ$, interpolation MAE $0.54 \pm 0.01^\circ$).

\begin{figure}[htbp]
\centering
\includegraphics[width=\linewidth]{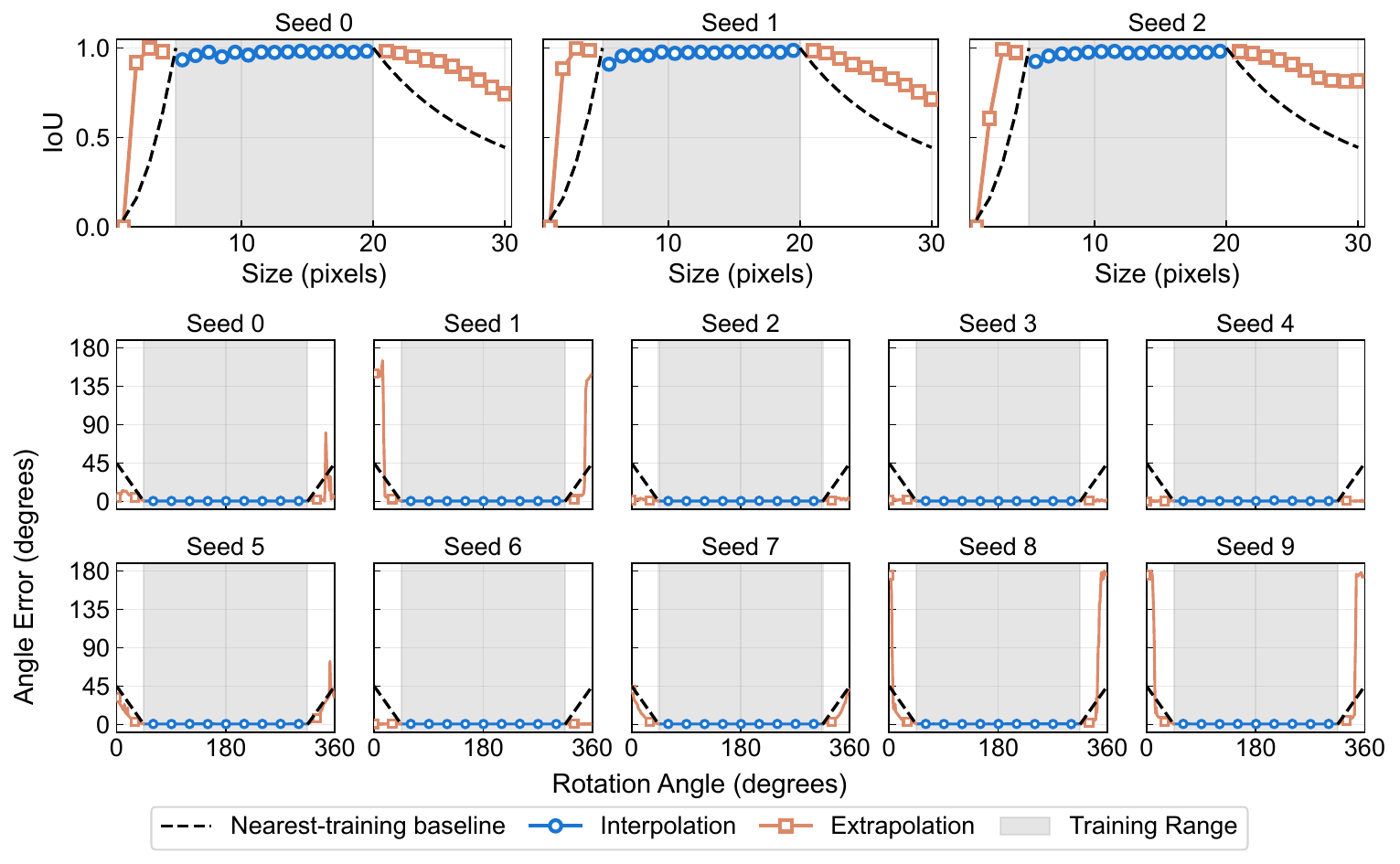}
\caption{\textbf{Per-seed \SkillSize\ and \SkillRotation\ curves.} Top. IoU per radius for each of the three \SkillSize\ seeds of \Cref{fig:radius_rotation}. Bottom. Angular error per angle for each of the ten \SkillRotation\ seeds. The dashed line and the shading are as in \Cref{fig:radius_rotation}.}
\label{fig:radius_rotation_seeds}
\end{figure}

\begin{figure}[htbp]
\centering
\includegraphics[width=\linewidth]{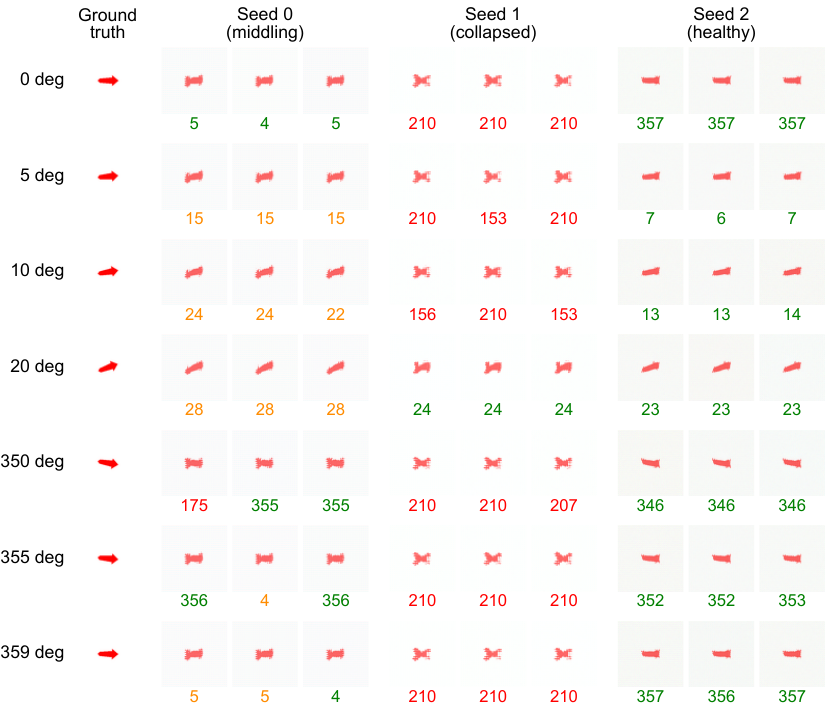}
\caption{\textbf{Rotation generations near $0^\circ$ from a middling, a collapsed, and a healthy seed.}
Each row is a target angle outside the training range, with three samples per seed. The number below each sample is its detected angle, colored by angular error (green ${\leq}5^\circ$, orange ${\leq}30^\circ$, red larger). Within ${\sim}15^\circ$ of $0^\circ$, the collapsed seed draws X-shaped blobs instead of arrows.}
\label{fig:rotation_3seed_qualitative}
\end{figure}

By extrapolation MAE, the ten seeds fall into three different behaviors. Four seeds (2, 3, 4, 6) generalize well across the full extrapolation grid (MAE $<2^\circ$, accuracy $99.9\%$ to $100\%$); three seeds (0, 5, 7) are middling ($8^\circ$ to $15^\circ$ MAE); and three seeds (1, 8, 9) collapse near $0^\circ$ ($43^\circ$ to $59^\circ$ MAE).
The middling and collapsed seeds average $45.7\%$ and $47.7\%$ extrapolation accuracy.
The collapse is concentrated within $\sim 15^\circ$ of $0^\circ$, the angle farthest from the training range; outside that region, accuracy is $99.8\%$ to $100\%$ for the four healthy seeds but at most $81.7\%$ for the other six.
The collapse rate is $3$ of $10$ ($30\%$, Wilson~\citep{wilson1927probable} $95\%$ CI: $[11\%, 60\%]$).
\Cref{fig:radius_rotation_seeds} plots the angular error per angle for every seed.
\Cref{fig:rotation_3seed_qualitative} shows generations from one seed per behavior.

\begin{table}[htbp]
\centering
\small
\renewcommand{\arraystretch}{1.05}
\begin{tabular}{@{}lcccccc@{}}
\toprule
& \multicolumn{2}{c}{\textbf{Train}} & \multicolumn{2}{c}{\textbf{Interp.}} & \multicolumn{2}{c}{\textbf{Extrap.}} \\
\cmidrule(lr){2-3} \cmidrule(lr){4-5} \cmidrule(lr){6-7}
\textbf{Seed} & MAE ($^\circ$) & Acc.\ @\,$5^\circ$ (\%) & MAE ($^\circ$) & Acc.\ @\,$5^\circ$ (\%) & MAE ($^\circ$) & Acc.\ @\,$5^\circ$ (\%) \\
\midrule
0 & 0.31 & 100.0 & 0.54 & 100.0 & 8.41 & 63.2 \\
1 & 0.30 & 100.0 & 0.53 & 100.0 & 48.94 & 53.2 \\
2 & 0.29 & 100.0 & 0.53 & 100.0 & 1.90 & 99.9 \\
3 & 0.27 & 100.0 & 0.53 & 100.0 & 1.49 & 100.0 \\
4 & 0.32 & 100.0 & 0.54 & 100.0 & 0.55 & 100.0 \\
5 & 0.37 & 100.0 & 0.57 & 100.0 & 15.33 & 32.3 \\
6 & 0.24 & 100.0 & 0.52 & 100.0 & 0.48 & 100.0 \\
7 & 0.30 & 100.0 & 0.53 & 100.0 & 12.41 & 41.6 \\
8 & 0.33 & 100.0 & 0.55 & 100.0 & 42.88 & 53.3 \\
9 & 0.28 & 100.0 & 0.53 & 100.0 & 59.28 & 36.6 \\
\midrule
mean${\pm}$std & $0.30_{\pm 0.03}$ & $100.0_{\pm 0.0}$ & $0.54_{\pm 0.01}$ & $100.0_{\pm 0.0}$ & $19.17_{\pm 22.46}$ & $68.0_{\pm 28.9}$ \\
\bottomrule
\end{tabular}
\vspace{0.35em}
\caption{\textbf{Per-seed rotation results for the main configuration.} The ten seeds share one setup (DiT-S/2, linear embedding, concatenation conditioning, no conditioning dropout) and the same training data ($[45^\circ, 315^\circ]$ at a $1^\circ$ step). They match inside the training range but split in extrapolation: seeds 2, 3, 4, and 6 are healthy, seeds 0, 5, and 7 are middling, and seeds 1, 8, and 9 collapse near $0^\circ$.}
\label{tab:rotation_per_seed_main}
\end{table}

\clearpage
\section{Additional Single-Skill Experiments}
\label{appendix:additional}
Apart from Appendix~\ref{appendix:matched_distance} and Appendix~\ref{appendix:dit_large_results}, which use the baseline and DiT-L models of Table~\ref{tab:single_property_results}, the \SkillCount\ models in this appendix train on radius-8 circles (radius 16 at $128\times128$) at random positions and on the same schedule as the other skills, rather than on the lattice and 3000 epochs of Appendix~\ref{appendix:count_geometry}.
In this setup, the $64\times64$ baseline fits $54.7\%$ of its training counts.

\subsection{Extrapolation at matched distance}
\label{appendix:matched_distance}
The out-of-range sets of Table~\ref{tab:single_property_results} reach different distances from the training range.
To compare skills at the same distance, we divide the distance outside the training range by the width of that range, $\delta = d_{\mathrm{out}} / \mathrm{width}$, and bin the baseline extrapolation accuracy by $\delta$ (Table~\ref{tab:matched_distance}).
For \SkillPosition, $d_{\mathrm{out}}$ is the $L_\infty$ distance to the training square, and the out-of-range positions form two rings at 4.6 and 5.8~px.
Every out-of-range angle and position has $\delta \leq 1/6$, while \SkillSize\ and \SkillCount\ are also tested farther out.
In this range, \SkillRotation\ is more accurate than \SkillPosition\ on average, and four of its ten seeds stay at or above $90\%$ on every value.
\SkillSize\ is at $100\%$ on every seed in the two bins it shares with \SkillRotation, and \SkillCount\ has no value with $\delta \leq 1/6$.

\begin{table}[htbp]
\centering
\small
\begin{tabular}{lcccc}
\toprule
 & \SkillSize & \SkillPosition & \SkillRotation & \SkillCount \\
\midrule
range width & 15~px & 36~px & $270^\circ$ & 5 \\
\midrule
$\delta \in [0, 0.07)$ & $100.0_{\pm 0.0}$ & \textemdash{} & $95.6_{\pm 7.6}$ & \textemdash{} \\
$\delta \in [0.07, 0.14)$ & $100.0_{\pm 0.0}$ & $48.8_{\pm 25.8}$ & $51.5_{\pm 43.9}$ & \textemdash{} \\
$\delta \in [0.14, 1/6]$ & \textemdash{} & $11.5_{\pm 15.7}$ & $43.7_{\pm 49.8}$ & \textemdash{} \\
$\delta \in (1/6, 0.25)$ & $72.5_{\pm 15.0}$ & \textemdash{} & \textemdash{} & $55.0_{\pm 49.9}$ \\
$\delta \in [0.25, 0.45)$ & $44.2_{\pm 23.1}$ & \textemdash{} & \textemdash{} & $50.0_{\pm 35.0}$ \\
$\delta \in [0.45, 2/3]$ & $0.0_{\pm 0.0}$ & \textemdash{} & \textemdash{} & \textemdash{} \\
\midrule
$\delta \leq 1/6$, mean & $100.0_{\pm 0.0}$ & $29.7_{\pm 16.8}$ & $68.0_{\pm 28.9}$ & \textemdash{} \\
$\delta \leq 1/6$, seeds ${\geq}\,90\%$ & 3/3 & 0/3 & 4/10 & \textemdash{} \\
\bottomrule
\end{tabular}
\vspace{0.35em}
\caption{\textbf{Extrapolation accuracy (\%) at matched distance from the training range.} $\delta$ is the distance outside the training range divided by the width of that range. Cells are mean$\pm$std over the seeds of Table~\ref{tab:single_property_results} (three, or ten for \SkillRotation), with the values in a bin weighted equally, and \textemdash{} marks bins with no tested value. The last two rows pool $\delta \leq 1/6$, which covers every out-of-range angle: the mean, and the number of seeds at or above $90\%$ on every value.}
\label{tab:matched_distance}
\end{table}

\subsection{Model size}
\label{appendix:dit_large_results}

The DiT-L row of \Cref{tab:single_property_results} scales the backbone from $22.20$M to ${\sim}306$M parameters and keeps the data, conditioning mechanism, optimizer, schedule, and sampler of DiT-S/2.
At the larger scale, training accuracy stays at $100\%$, and mean extrapolation accuracy on \SkillSize\ and \SkillPosition\ is below $50\%$.
Scaling to DiT-L improves \SkillCount\ and \SkillPosition\ on average ($+19.6$~pp and $+12.8$~pp on extrapolation, with a large spread across seeds), but does \emph{not} improve \SkillSize\ ($-7.0$~pp) or \SkillRotation\ ($-6.0$~pp).
On \SkillRotation, the three DiT-L seeds reach $52\%$ to $80\%$ extrapolation accuracy, and one of them collapses near $0^\circ$, as three of the ten DiT-S/2 seeds do (Appendix~\ref{appendix:rotation_ablations}).

\subsection{Dataset size}
\label{appendix:data_scaling}
Table~\ref{tab:data_scaling} reports the baseline trained on nested subsets of 10k, 20k, 40k, and 80k images for each skill, three seeds each, at a budget of 78k to 80k optimizer steps (1000 epochs at 10k images, 125 at 80k).
Extrapolation here is scored on a narrower out-of-range grid than Table~\ref{tab:single_property_results} (radii up to 25 rather than 30; positions from the training boundary rather than 4~px beyond it), so absolute \SkillSize\ and \SkillPosition\ values are higher than in the main table; compare within a row.
No skill improves with dataset size beyond seed noise.

\begin{table}[htbp]
\centering
\small
\begin{tabular}{lcccc}
\toprule
skill & 10k & 20k & 40k & 80k \\
\midrule
\SkillSize & $77.6_{\pm 6.4}$ & $76.3_{\pm 11.7}$ & $75.4_{\pm 10.0}$ & $80.4_{\pm 2.5}$ \\
\SkillPosition & $67.7_{\pm 20.1}$ & $69.1_{\pm 13.9}$ & $70.1_{\pm 9.5}$ & $73.5_{\pm 7.6}$ \\
\SkillRotation & $81.2_{\pm 32.5}$ & $79.3_{\pm 34.4}$ & $87.8_{\pm 21.2}$ & $64.2_{\pm 31.1}$ \\
\SkillCount & $14.6_{\pm 2.9}$ & $17.5_{\pm 6.5}$ & $16.3_{\pm 9.4}$ & $13.3_{\pm 3.8}$ \\
\bottomrule
\end{tabular}
\vspace{0.35em}
\caption{\textbf{Extrapolation accuracy (\%) against training-set size.} Mean$\pm$std over three seeds.}
\label{tab:data_scaling}
\end{table}

\subsection{Sequence length and latent space}
\label{appendix:token_matched}
The VAE row of Table~\ref{tab:single_property_results} compresses a $64\times64$ image to an $8\times8$ latent, which DiT-S/2 turns into 16 tokens, while the pixel baseline uses 1024.
Table~\ref{tab:token_matched} holds parameters at ${\sim}22$M and varies the token count by changing the patch size.
A latent model with 64 tokens (patch size 1) reaches $36.1\%$ on \SkillSize\ extrapolation, far above the 16-token latent model, but only $18.5\%$ on \SkillRotation, no better than the 16-token latent model ($21.2\%$); a pixel model with 16 tokens (patch size 16) reaches $2.7\%$ on \SkillRotation\ and fits its training data poorly, with each token encoding a $16\times16$ patch (768 values).

\begin{table}[htbp]
\centering
\small
\begin{tabular}{llccc}
\toprule
space & model & tokens & \SkillSize & \SkillRotation \\
\midrule
latent & DiT-S/2 & 16 & 7.1 & 21.2 \\
pixel & DiT-S/16 & 16 & \textemdash & 2.7 \\
latent & DiT-S/1 & 64 & 36.1 & 18.5 \\
pixel & DiT-S/2 & 1024 & 51.5 & 68.0 \\
\bottomrule
\end{tabular}
\vspace{0.35em}
\caption{\textbf{Extrapolation accuracy (\%) as the token count varies at about 22M parameters.} The DiT-S/2 rows repeat the means from Table~\ref{tab:single_property_results}. The other rows are one seed each, scored on the same out-of-range values as that table.}
\label{tab:token_matched}
\end{table}

\subsection{Resolution}
\label{appendix:hires}
Table~\ref{tab:hires} reports the baseline at $128\times128$ in pixel space (DiT-S/4, three seeds) and in a VAE latent space (one seed), with length thresholds scaled by two and the count blob threshold by four.
Each split is capped at 100 evenly spaced conditions, \SkillSize\ extrapolation is scored on radii 2 to 9 and 41 to 60, and \SkillPosition\ extrapolation starts at the training boundary.
\SkillSize\ extrapolation is similar to $64\times64$ ($46.8\%$ vs.\ $51.5\%$, within the seed spread); \SkillPosition\ extrapolates further in absolute pixels but about as far relative to image size; \SkillRotation\ seeds again fall into different behaviors as at $64\times64$. In pixel space, \SkillCount\ fits only $36.4\%$ of its training counts, against $54.7\%$ at $64\times64$.
The latent model (64 tokens) extrapolates \SkillSize\ and \SkillPosition\ about as well as the pixel models, and its \SkillRotation\ accuracy ($13.7\%$) is close to that of the weakest pixel seed ($14.3\%$).

\begin{table}[htbp]
\centering
\small
\begin{tabular}{lcccccc}
\toprule
 & \multicolumn{3}{c}{pixel} & \multicolumn{3}{c}{VAE latent} \\
\cmidrule(lr){2-4} \cmidrule(lr){5-7}
skill & train & interp & extra & train & interp & extra \\
\midrule
\SkillSize & $100.0_{\pm 0.0}$ & $100.0_{\pm 0.0}$ & $46.8_{\pm 6.6}$ & $100.0$ & $90.3$ & $47.5$ \\
\SkillPosition & $100.0_{\pm 0.0}$ & $100.0_{\pm 0.0}$ & $73.3_{\pm 11.8}$ & $100.0$ & $100.0$ & $72.2$ \\
\SkillRotation & $100.0_{\pm 0.0}$ & $100.0_{\pm 0.0}$ & $59.6_{\pm 43.1}$ & $100.0$ & $100.0$ & $13.7$ \\
\SkillCount & $36.4_{\pm 4.9}$ & \textemdash{} & $13.8_{\pm 3.3}$ & $89.2$ & \textemdash{} & $55.0$ \\
\bottomrule
\end{tabular}
\vspace{0.35em}
\caption{\textbf{Accuracy (\%) at $128\times128$.} Pixel: mean$\pm$std over three seeds. Latent: one seed.}
\label{tab:hires}
\end{table}

\subsection{Text-encoder conditioning}
\label{appendix:text_cond}
Table~\ref{tab:text_cond} replaces the numeric embedding with a frozen CLIP ViT-B/32 text encoder~\citep{radford2021learning}: each skill value selects the embedding of a sentence that states it, and a learned linear layer takes the place of $\mathbf{W}v + \mathbf{b}$ (Appendix~\ref{appendix:architecture}), with the rest of the model trained as usual.
Training accuracy matches numeric conditioning on all three skills, but interpolation falls from $96.8\%$ to $13.3\%$ for \SkillSize\ and from $100.0\%$ to $60.9\%$ for \SkillRotation, and extrapolation collapses, with \SkillRotation\ dropping from $68.0\%$ to $8.4\%$.
This suggests that the ordered numeric input matters for extrapolation.

\begin{table}[htbp]
\centering
\small
\begin{tabular}{lccc}
\toprule
skill & train & interp & extra \\
\midrule
\SkillSize & $100.0$ & $13.3$ & $0.0$ \\
\SkillRotation & $100.0$ & $60.9$ & $8.4$ \\
\SkillCount & $54.2$ & \textemdash{} & $1.3$ \\
\bottomrule
\end{tabular}
\vspace{0.35em}
\caption{\textbf{Accuracy (\%) with CLIP text conditioning in place of the numeric embedding.} One seed.}
\label{tab:text_cond}
\end{table}

\subsection{Textured renders}
\label{appendix:realistic}
Table~\ref{tab:realistic} reports the baseline trained on renders with textured backgrounds, radial shading, and varied object color.
Otsu thresholding fails on textured backgrounds, so these runs are scored with a saturation-mask extractor.
On the textured renders, \SkillSize\ extrapolation is about 20 points lower and \SkillRotation\ about 20 points higher than on the simple renders, although the \SkillRotation\ gain is within its seed spread.
\SkillRotation\ extrapolates best on these renders and \SkillCount\ worst.

\begin{table}[htbp]
\centering
\small
\begin{tabular}{lccc}
\toprule
skill & simple & textured & $\Delta$ \\
\midrule
\SkillSize & 77.3 & 56.9 & $-20.4$ \\
\SkillPosition & 66.5 & 67.7 & +1.2 \\
\SkillRotation & 66.0 & 86.3 & +20.3 \\
\SkillCount & 12.9 & 15.4 & +2.5 \\
\bottomrule
\end{tabular}
\vspace{0.35em}
\caption{\textbf{Extrapolation accuracy (\%) on simple and textured renders.} Both columns are scored on the narrower out-of-range grid of Appendix~\ref{appendix:data_scaling}.}
\label{tab:realistic}
\end{table}

\subsection{Pre-trained text-to-image models}
\label{appendix:t2i}

We prompt eight pre-trained text-to-image models for the four numeric skills and score the outputs with rule-based extractors adapted to these images: a redness mask, watershed counting, and the grid cell, equivalent-circle diameter, and arrow direction of the largest red blob that neither touches the image border nor encloses a hole, or of the largest blob if none qualifies (Table~\ref{tab:t2i}).
Prompts request red circles on a white background with a stated count ($1$ to $10$), position (nine grid cells), size (six diameters from $10\%$ to $90\%$ of the frame width), or rotation (eight compass directions for an arrow).
A size passes within $25\%$ of the requested diameter and an angle within $22.5^\circ$, while count and position must match exactly.
Accuracy is computed over the $97.9\%$ of samples that contain a measurable red subject, and bold marks accuracy above chance level, which we take as the 95th percentile of accuracy when the requests are randomly shuffled across images.

\begin{figure}[htbp]
\centering
\includegraphics[width=0.78\linewidth]{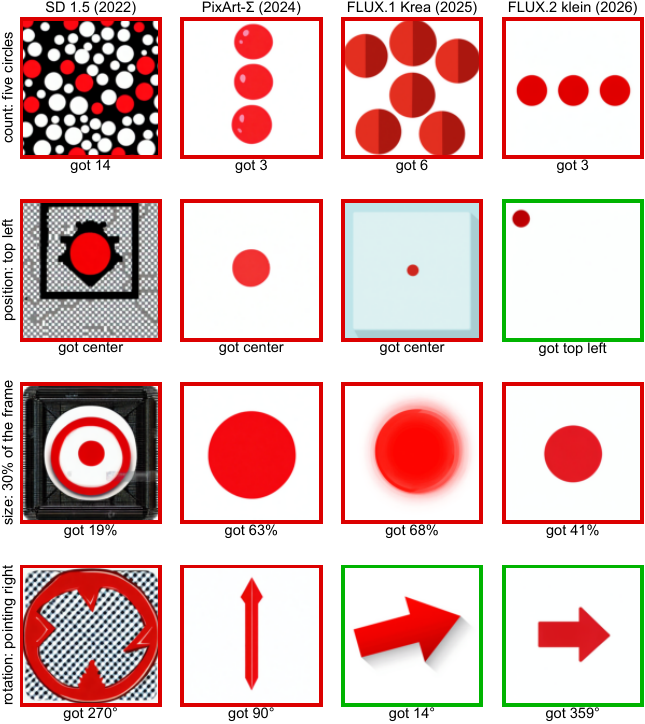}
\caption{\textbf{Median-error samples from pre-trained text-to-image models for one prompt per skill.}
Each tile is the sample with the median error among 20 generations. The value measured by the Table~\ref{tab:t2i} scorer is printed below it, and a green border marks a sample that counts as correct in that table.
In these samples, only FLUX.2 klein places the object, the two FLUX models orient it, and no model gets the count or size right.}
\label{fig:t2i_examples}
\end{figure}

\Cref{fig:t2i_examples} shows the median-error sample for one prompt per skill from four of the eight models.
No model is above chance level on \SkillSize; \SkillCount\ tops out at $58\%$; \SkillPosition\ improves sharply only in Shuttle 3 and FLUX.2 klein, and \SkillRotation\ most in FLUX.2 klein.
For \SkillSize, regressing the measured diameter on the requested one gives slopes from $0.013$ to $0.076$ across the eight models, against $1$ for a model that follows the request.
These numbers compare models under one protocol and are not comparable to Table~\ref{tab:single_property_results}. The prompted value ranges and the tolerances differ from VGL's, and the models were not trained on this distribution.

\begin{table}[htbp]
\centering
\small
\begin{tabular}{lcccc}
\toprule
model & \SkillCount & \SkillPosition & \SkillSize & \SkillRotation \\
\midrule
Stable Diffusion 1.5~\citep{rombach2022high} & \textbf{8.6} & 11.5 & 24.6 & 16.4 \\
SDXL~\citep{podell2024sdxl} & 1.0 & \textbf{15.7} & 11.4 & 13.1 \\
PixArt-$\alpha$~\citep{chen2024pixartalpha} & \textbf{32.0} & 11.1 & 31.7 & 13.8 \\
PixArt-$\Sigma$~\citep{chen2024pixartsigma} & \textbf{32.5} & 11.1 & 26.7 & 13.8 \\
HunyuanDiT~\citep{li2024hunyuandit} & \textbf{30.0} & 11.1 & 22.8 & \textbf{18.1} \\
Shuttle 3~\citep{shuttle3diffusion} & \textbf{49.2} & \textbf{61.2} & 16.7 & \textbf{31.7} \\
FLUX.1 Krea~\citep{flux1kreadev} & \textbf{41.0} & \textbf{32.2} & 20.0 & \textbf{35.6} \\
FLUX.2 klein~\citep{flux2klein} & \textbf{58.0} & \textbf{89.4} & 16.7 & \textbf{74.4} \\
\bottomrule
\end{tabular}
\vspace{0.35em}
\caption{\textbf{Skill accuracy (\%) of pre-trained text-to-image models under one prompting protocol.} \textbf{Bold} marks accuracy above chance level.}
\label{tab:t2i}
\end{table}

\clearpage
\section{Qualitative Analysis of Generalization}
\label{appendix:qualitative_analysis}

\begin{figure}[htbp]
\centering
\includegraphics[width=0.92\linewidth]{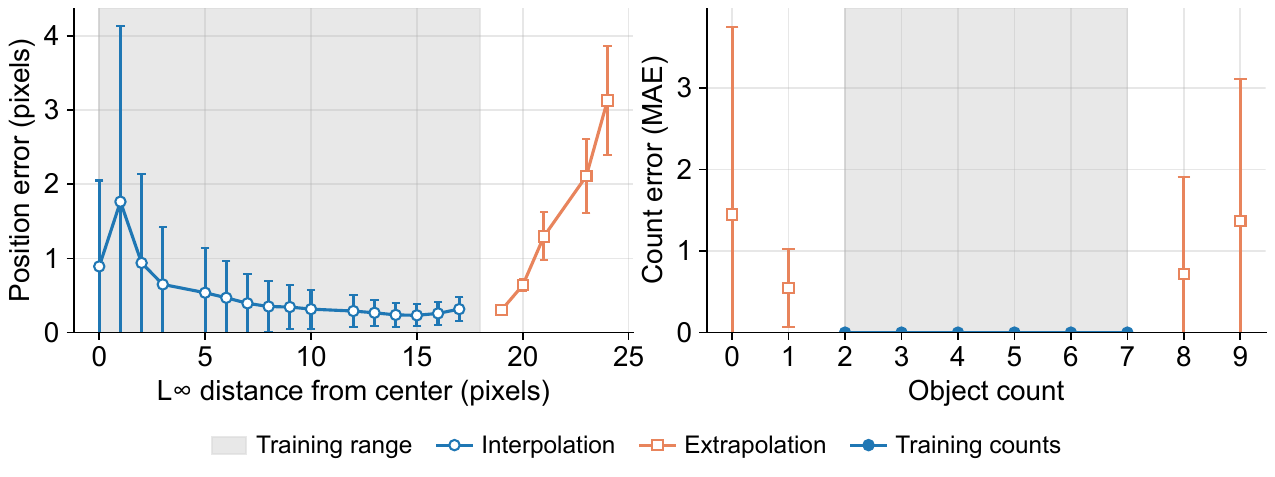}
\caption{\textbf{Position and count errors stay low inside the training range and grow outside it.}
Left. Mean position error of the \SkillPosition\ baseline against $L_\infty$ distance from the image center.
Right. Mean absolute count error of the \SkillCount\ baseline.
Error bars show one standard deviation over three seeds.
The wide bar at count 0 reflects one seed that draws about four objects while the other two mostly leave the frame empty.}
\label{fig:position_count}

\end{figure}

We provide qualitative analysis of model behavior across all four skills.
\Cref{fig:qualitative} summarizes per-skill generalization, with borders marking success or failure.
\Cref{fig:position_count} summarizes generalization behavior for \SkillPosition\ and \SkillCount.
The \SkillSize\ and \SkillRotation\ curves are in Figure~\ref{fig:radius_rotation}, with \SkillSize\ extrapolation attempts in Figure~\ref{fig:size_failures} and per-seed \SkillRotation\ results in \Cref{appendix:rotation_ablations}.
\Cref{fig:count_failures} provides additional qualitative failure modes for \SkillCount\ beyond the curve view.

\paragraph{Size extrapolation limits.}
The size model achieves strong interpolation but fails at extreme extrapolation (\Cref{fig:size_failures}).
Across the three seeds of \Cref{tab:single_property_results}, radius 24 always succeeds (at least half of the samples pass), radius 25 succeeds for two seeds, radius 26 for one, and radii 27 to 30 fail for all.
Radius 2 succeeds for one seed ($45.0\%$ of samples over the three seeds), and for radius 1 pixel the model fails to generate anything.
For larger sizes beyond training, circles lose their shape and appear more square-like.
Single runs trained on radii 9 to 16, 5 to 20, and 11 to 26 succeed from radius 5 to 18, 3 to 24, and 7 to at least 32, respectively, so the extrapolation limits move with the training range rather than staying at fixed sizes.

\paragraph{Position extrapolation.}
The position model demonstrates good interpolation and attempts extrapolation to far corners (\Cref{fig:position_failures}).
However, at extreme positions, the generated circle is pulled inward, sometimes with a small extra fragment.

\paragraph{Count challenges.}
The count model trained on $\{2, 3, 4, 5, 6, 7\}$ circles reproduces every trained count exactly but extrapolates poorly (\Cref{fig:position_count,fig:count_failures}).
On out-of-range queries, the model shown in \Cref{fig:qualitative,fig:count_failures} produces the correct count for $1$, $8$, and $9$ most of the time and almost never for $0$, where it draws one to three objects.
One of the three seeds in \Cref{tab:single_property_results} fails on every out-of-range count except $0$, where it leaves the frame empty, so the three-seed extrapolation accuracy is $52.5\%$.
Because the model fits every training count, these failures are not due to underfitting.

\begin{figure}[htbp]
\centering
\begin{subfigure}[t]{0.48\linewidth}
  \centering
  \includegraphics[width=\linewidth]{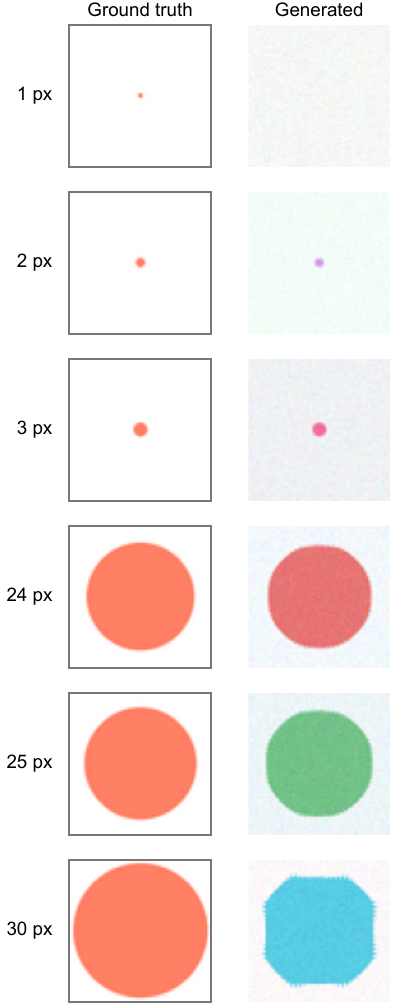}
  \caption{\SkillSize\ extrapolation attempts.}
  \label{fig:size_failures}
\end{subfigure}
\hfill
\begin{subfigure}[t]{0.48\linewidth}
  \centering
  \includegraphics[width=\linewidth]{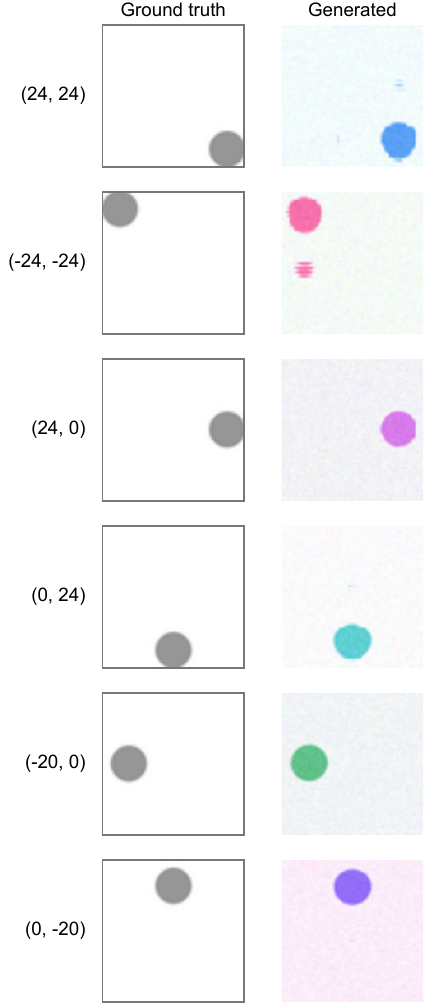}
  \caption{\SkillPosition\ extrapolation attempts.}
  \label{fig:position_failures}
\end{subfigure}
\caption{\textbf{Beyond the training range, large circles turn square-like and distant circles are pulled inward.}
Rows are labeled by radius or by $(x, y)$ offset from the image center, and each pairs the ground truth (canvas outlined) with a generated sample.
At the farthest offsets, the generated circle usually lands just outside the training range, sometimes with a small extra fragment.
}
\label{fig:qualitative_failures}
\end{figure}

\begin{figure}[htbp]
\centering
\includegraphics[width=0.95\linewidth]{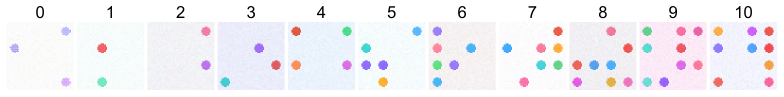}
\caption{\textbf{Count extrapolation failure modes.}
Each panel shows one generated sample for the requested count, from a model trained on 2 to 7 objects.
Counts 1, 8, and 9 are usually correct, although the sample shown for 1 has two objects.}
\label{fig:count_failures}
\end{figure}

\clearpage
\section{Compositional Results}
\label{appendix:compositional}

\paragraph{Analysis of \SkillSize~+~\SkillShape\ failure.}
\label{appendix:size_shape_failure}
The \SkillSize\ and \SkillShape\ composition difficulties may stem from the context-dependent nature of the radius parameter. A circle with radius 10 differs visually from a square or triangle with the same radius parameter. This mismatch means the same numeric size value corresponds to different pixel footprints across shapes, which can make held-out pairings harder to learn (training combinations score 100\%).
\SkillSize\ is the weaker skill of this pair. At 75\% coverage, \SkillSize\ reaches 72.3\% against 93.7\% for \SkillShape, and most \SkillSize\ failures draw the requested shape at the wrong size (22\% of samples, against 6\% that get both wrong). Two thirds of the \SkillSize\ misses are radius-12 triangles drawn about 8\% too large.

\paragraph{Analysis of \SkillShape~+~\SkillCount\ failure.}
\SkillShape~+~\SkillCount\ is the one pair in Table~\ref{tab:compositional_results} whose interpolation fails for both skills at 75\% coverage (10.0\% shape accuracy, 46.4\% count accuracy). In 79\% of its samples, touching objects merge into one component that matches no shape and must be split before counting.

\paragraph{Combinations per attribute value.}
\label{appendix:k_analysis}

Table~\ref{tab:k_analysis} lists twelve \SkillColor~+~\SkillShape\ cells from a vocabulary-size sweep ($S \in \{4, 8\}$ shapes, $C \in \{8, 16\}$ colors, coverage $25$ to $75\%$, about $15{,}000$ training images in every cell) with $k = K/S$, where $K$ is the number of training combinations.
The eight-shape vocabulary adds pentagon, hexagon, star, and cross to the shapes of Table~\ref{tab:skill_domains}, and the 16-color vocabulary adds lime, teal, pink, brown, navy, maroon, olive, and azure to its colors.
Each image shows one centered object of radius 12 px, and a sample counts as correct only when both its \SkillShape\ and its \SkillColor\ are right under the metrics of Appendix~\ref{appendix:eval_protocol}.
Across the twelve cells, $\log_2 k$ alone gives $R^2 = 0.62$ on held-out-combination accuracy, against $0.60$ for $\log_2 K$ and $0.49$ for coverage; with $n = 12$ the first two are not separable, but every cell with $k \geq 6$ reaches $100\%$.
Five random draws of the training combinations at $S = 4$, $C = 8$, and $50\%$ coverage, each trained with one seed, reach $100.0$, $100.0$, $100.0$, $100.0$, and $87.5\%$ on held-out combinations, within the range of the three training seeds of that cell ($68.8$ to $100\%$).
For the same $S$ and $C$, coverage moves held-out accuracy from $38.4\%$ at $25\%$ to $100\%$ at $75\%$, so which combinations are held out matters less than how many.

\begin{table}[htbp]
\centering
\small
\begin{tabular}{lcccccc}
\toprule
$S$ & $C$ & coverage & $K$ & $k$ & seeds & held-out (\%) \\
\midrule
4 & 8 & 25\% & 8 & 2 & 3 & $38.4_{\pm 38.4}$ \\
8 & 8 & 25\% & 16 & 2 & 3 & $42.0_{\pm 25.2}$ \\
4 & 16 & 25\% & 16 & 4 & 3 & $59.1_{\pm 1.9}$ \\
8 & 16 & 25\% & 32 & 4 & 1 & 100.0 \\
4 & 8 & 50\% & 16 & 4 & 3 & $89.6_{\pm 18.0}$ \\
8 & 8 & 50\% & 32 & 4 & 3 & $100.0_{\pm 0.0}$ \\
4 & 8 & 75\% & 24 & 6 & 3 & $100.0_{\pm 0.0}$ \\
8 & 8 & 75\% & 48 & 6 & 3 & $100.0_{\pm 0.0}$ \\
4 & 16 & 50\% & 32 & 8 & 1 & 100.0 \\
8 & 16 & 50\% & 64 & 8 & 1 & 100.0 \\
4 & 16 & 75\% & 48 & 12 & 1 & 100.0 \\
8 & 16 & 75\% & 96 & 12 & 1 & 100.0 \\
\bottomrule
\end{tabular}
\vspace{0.35em}
\caption{\textbf{Held-out \SkillColor~+~\SkillShape\ accuracy saturates once $k$, the average number of training combinations per shape, reaches about six.}
$S$, $C$, and $K$ are the numbers of shapes, colors, and training combinations, so $k = K/S$.
Every cell trains on about $15{,}000$ images, and a held-out sample counts only when both skills are correct. Three-seed rows give the mean with the standard deviation as a subscript.}
\label{tab:k_analysis}
\end{table}

\paragraph{Out-of-range splits at lower coverage.}
Section~\ref{sec:distribution_learning} uses the $75\%$ models of \Cref{tab:compositional_results}, which reach near $100\%$ interpolation accuracy.
The coupling of Section~\ref{sec:distribution_learning} also holds at $50\%$ coverage: an out-of-range \SkillPosition\ lowers \SkillSize\ accuracy from $96.0\%$ to $20.8\%$ and \SkillRotation\ accuracy from $73.8\%$ to $57.9\%$.
At $25\%$, an out-of-range \SkillPosition\ does not lower \SkillSize\ ($46.3\%$ in range, $49.5\%$ out of range), and \SkillRotation\ interpolation is only $5.6\%$, so the $39.6\%$ on its out-of-range split says little about interdependence.

\paragraph{Out-of-range rotation.}
On the \CatC{\SkillRotation} split of Table~\ref{tab:compositional_results}, the \SkillPosition~+~\SkillRotation\ models draw nearly every arrow at the nearest training angle. Unlike the single-skill \SkillRotation\ models, they are trained on four angles $90^\circ$ apart and embed the angle without periodicity (Appendix~\ref{appendix:architecture}).

\clearpage
\section{Three- and Four-Skill Models}
\label{appendix:multi_prop}

We evaluate trained three-skill and four-skill compositional models (DiT-S/2, $75\%$ coverage, two training runs for the three-skill setting and one for the four-skill setting) on splits built from their training grid. The grid has radii 6, 8, \dots, 14, four shapes, and, for the four-skill model, eight colors. Its positions lie on a 7-pixel grid from $-14$ to $14$ in $x$ and $y$, of which the 20 with $x \leq 7$ are used in training. The splits are training combinations (\CatA), held-out in-range combinations (\CatB), radius 16 or 18 (\CatC, \SkillSize\ OOD), $x = 14$ (\CatC, \SkillPosition\ OOD), and both (\CatD).
\Cref{tab:multi_prop_interdependence} reports the per-sample accuracy of every skill, with 20 samples per combination and the multi-skill metrics of Appendix~\ref{appendix:eval_protocol}.

Section~\ref{sec:distribution_learning} summarizes the four-skill model; \Cref{tab:multi_prop_interdependence} adds the two three-skill runs and the splits where both \SkillSize\ and \SkillPosition\ are out of range.
In both three-skill runs, an out-of-range \SkillPosition\ also lowers \SkillShape\ ($20.4\%$ and $59.9\%$) and \SkillSize\ ($30.9\%$ and $84.7\%$), while an out-of-range \SkillSize\ leaves \SkillShape\ and \SkillPosition\ at $100\%$.
In the first three-skill run and the four-skill model, most samples with only \SkillPosition\ out of range contain a malformed object that matches no shape ($55\%$ and $65\%$, against $13\%$ in the second three-skill run).
The two-skill out-of-range splits not listed in Table~\ref{tab:compositional_results} show the same pattern at 75\% coverage.
With \SkillPosition\ out of range, \SkillShape\ falls from $100.0\%$ to $33.0\%$ in \SkillPosition~+~\SkillShape, while \SkillColor\ stays at $99.9\%$ in \SkillColor~+~\SkillPosition.
With \SkillSize\ or \SkillCount\ out of range, the other skill does not fall: \SkillShape\ is at $99.4\%$ in \SkillSize~+~\SkillShape\ ($93.7\%$ in range), \SkillColor\ at $100.0\%$ in \SkillColor~+~\SkillCount, and \SkillShape\ at $51.5\%$ in \SkillShape~+~\SkillCount\ ($10.0\%$ in range).

\begin{table*}[htbp]
\centering
\footnotesize
\setlength{\tabcolsep}{4pt}
\renewcommand{\arraystretch}{1.05}
\begin{tabular}{@{}llcccc@{}}
\toprule
\textbf{Model} & \textbf{Split} & \textbf{\SkillSize\ (\%)} & \textbf{\SkillPosition\ (\%)} & \textbf{\SkillShape\ (\%)} & \textbf{\SkillColor\ (\%)} \\
\midrule
\multirow{5}{*}{3-skill, run 1}
 & \CatA\ (training)           & 100.0 & 100.0 & 100.0 & \NA \\
 & \CatB\ (in-range)           & 99.6 & 100.0 & 100.0 & \NA \\
 & \CatC\ (\SkillSize\ OOD)    & \fail{37.6} & 100.0 & 100.0 & \NA \\
 & \CatC\ (\SkillPosition\ OOD) & \fail{30.9} & \fail{0.0} & \fail{20.4} & \NA \\
 & \CatD\ (both OOD)           & \fail{6.6} & \fail{0.0} & \fail{25.5} & \NA \\
\midrule
\multirow{5}{*}{3-skill, run 2}
 & \CatA\ (training)           & 100.0 & 100.0 & 100.0 & \NA \\
 & \CatB\ (in-range)           & 97.9 & 100.0 & 100.0 & \NA \\
 & \CatC\ (\SkillSize\ OOD)    & \fail{37.2} & 100.0 & 100.0 & \NA \\
 & \CatC\ (\SkillPosition\ OOD) & \fail{84.7} & \fail{0.0} & \fail{59.9} & \NA \\
 & \CatD\ (both OOD)           & \fail{44.6} & \fail{0.0} & \fail{56.5} & \NA \\
\midrule
\multirow{5}{*}{4-skill}
 & \CatA\ (training)           & 100.0 & 100.0 & 100.0 & 100.0 \\
 & \CatB\ (in-range)           & 100.0 & 100.0 & 100.0 & 100.0 \\
 & \CatC\ (\SkillSize\ OOD)    & \fail{54.6} & 100.0 & 100.0 & 100.0 \\
 & \CatC\ (\SkillPosition\ OOD) & \fail{79.5} & \fail{0.0} & \fail{32.0} & 100.0 \\
 & \CatD\ (both OOD)           & \fail{50.5} & \fail{0.0} & \fail{19.2} & 100.0 \\
\bottomrule
\end{tabular}
\vspace{0.35em}
\caption{\textbf{An out-of-range \SkillPosition\ degrades \SkillSize\ and \SkillShape\ in every model, while an out-of-range \SkillSize\ leaves the other skills intact.}
The three-skill models condition on \SkillSize, \SkillPosition, and \SkillShape, and the four-skill model adds \SkillColor. All are trained at $75\%$ coverage.
Entries are per-skill accuracy (\%). \fail{Gray} marks accuracy below $90\%$.}
\label{tab:multi_prop_interdependence}
\end{table*}

\clearpage
\section{Embedding Manifold Analysis}
\label{appendix:embedding_manifold}

\begin{figure}[htbp]
\centering
\includegraphics[width=0.95\linewidth]{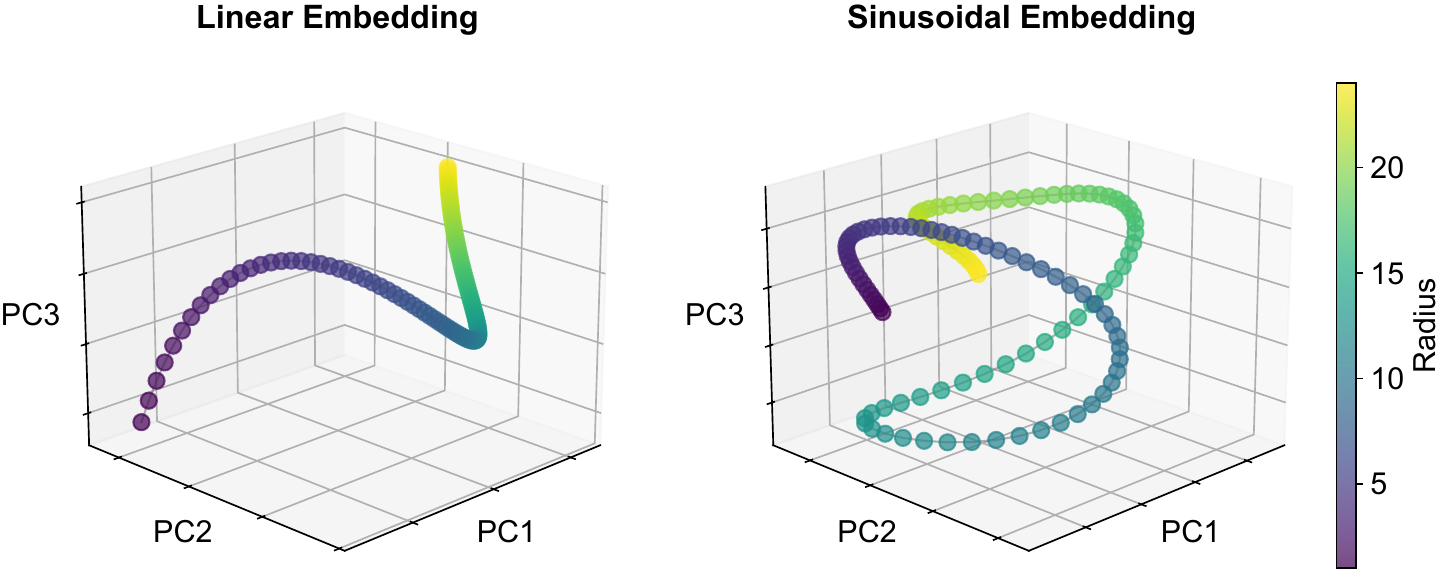}
\caption{\textbf{The linear \SkillSize\ embedding follows a near-monotone path, while the sinusoidal one follows a longer, winding path.}
Each point is the 3D PCA projection of the learned conditioning embedding for one radius from $1$ to $24$ pixels, colored by radius. Both models are DiT-S/2 trained on radii $5$ to $20$.}
\label{fig:manifold_comparison}
\end{figure}

To compare how different conditioning embedding schemes organize skill information, we visualize learned embedding manifolds for \SkillSize\ conditioning.
\Cref{fig:manifold_comparison} shows 3D PCA projections of the conditioning token vector at radii from $1$ to $24$ pixels in steps of $0.2$ (the embedding fed into the transformer after the learnable projection to hidden dimension $384$).
The linear embedding traces a smooth, near-monotone trajectory in PCA space, while the sinusoidal embedding follows a longer, winding path, consistent with its multi-frequency construction.
The figure is a qualitative illustration of why the linear embedding's monotone structure may be easier to extrapolate from.

\clearpage
\section{Predicting Extrapolation Accuracy from Per-Skill Features}
\label{appendix:mechanistic}

\paragraph{Setup.}
We summarize the three-seed ablation sweep from \Cref{tab:single_property_results} as a small tabular dataset with one row per skill, variant, and seed.
This gives $n{=}103$ rows from four skills, eight variants, and three seeds (ten for the \SkillRotation\ baseline).
The response is extrapolation accuracy \texttt{extra\_acc}.
The predictors are the skill, embedding type, conditioning mechanism, pixel or VAE latent space, DDPM or flow objective, and architecture (DiT, including DiT-L, or U-Net).
We fit ordinary least-squares models in scikit-learn~\citep{sklearn} with one-hot encoded categorical features.
The dropped reference designs are linear embedding, concatenation conditioning, pixel space, DDPM, DiT, and rotation, so each coefficient is a percentage point change relative to that reference design.

\begin{figure}[htbp]
\centering
\includegraphics[width=0.95\linewidth]{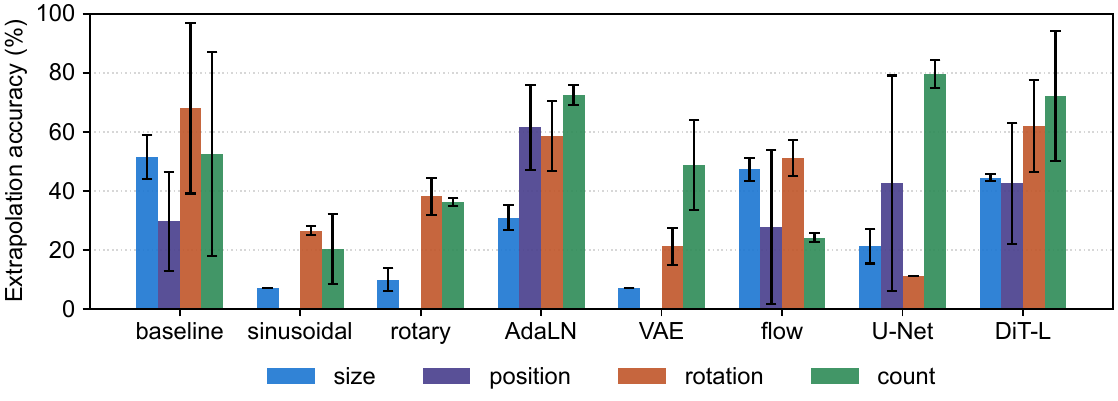}
\caption{\textbf{The same design choice can help one skill extrapolate and hurt another.}
Bars give extrapolation accuracy for each skill and variant, averaged over three seeds (ten for the \SkillRotation\ baseline), and error bars show one standard deviation. AdaLN, for example, helps \SkillPosition\ but hurts \SkillSize.
\Cref{fig:mechanistic_variants} summarizes these effects as per-skill regression coefficients.}
\label{fig:mechanistic_extra}
\end{figure}

\begin{figure}[htbp]
\centering
\begin{subfigure}[b]{0.45\linewidth}
\centering
\includegraphics[width=\linewidth]{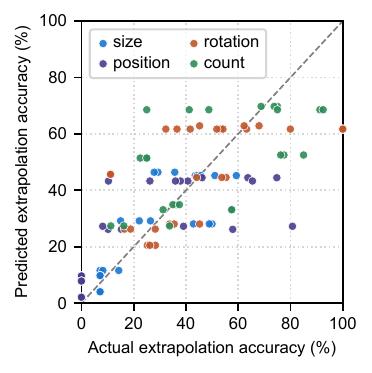}
\caption{Predicted vs.\ actual extrapolation accuracy.}
\label{fig:mechanistic_pred}
\end{subfigure}\hfill
\begin{subfigure}[b]{0.53\linewidth}
\centering
\includegraphics[width=\linewidth]{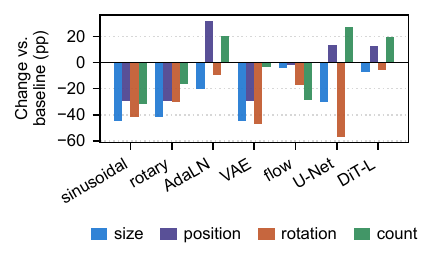}
\caption{Per-skill variant effects.}
\label{fig:mechanistic_variants}
\end{subfigure}
\caption{\textbf{One regression across all skills leaves large residuals, partly because a design choice can push different skills in opposite directions.}
Panel~(\subref{fig:mechanistic_pred}) plots that fit's predictions from skill and design choices against measured extrapolation accuracy, with one point per skill, variant, and seed.
Panel~(\subref{fig:mechanistic_variants}) shows the coefficients of separate per-skill fits, as percentage-point changes from the baseline.}
\label{fig:mechanistic_combined}
\end{figure}

\paragraph{Regression results.}
The across-skill categorical model attains $R^2 = 0.540$.
Adding interpolation accuracy as a predictor leaves the fit essentially unchanged, with $R^2 = 0.553$ and a coefficient of $0.712$ (standard error $0.43$) on \texttt{interp\_acc}.
A model using only skill, training accuracy, and interpolation accuracy as predictors explains substantially less variance ($R^2 = 0.183$).
\Cref{fig:mechanistic_extra} breaks extrapolation accuracy down by skill and variant, and \Cref{fig:mechanistic_pred} shows that the categorical model captures broad trends but leaves substantial residuals, especially for \SkillRotation, where the failure near $0^\circ$ is seed driven.
The largest across-skill coefficients are sinusoidal embedding ($-41.2$), VAE latent space ($-35.4$), and rotary embedding ($-33.6$).
Fitting each skill separately, with one indicator per variant and baseline as the reference, gives $R^2 = 0.96$ for \SkillSize, $0.72$ for \SkillCount, $0.66$ for \SkillPosition, and $0.60$ for \SkillRotation. Each value is the share of that skill's variance that lies between variants.

\paragraph{Interpretation and caveats.}
First, in-range performance is a weak proxy for extrapolation.
Within-skill Spearman correlations between interpolation and extrapolation accuracy are weak for \SkillSize\ and \SkillPosition\ ($|\rho| \leq 0.30$). They are undefined for \SkillRotation, whose interpolation accuracy is $100\%$ for every variant, and for \SkillCount, which has no interpolation split.
Second, design effects are skill dependent.
For example, AdaLN improves \SkillPosition\ but hurts \SkillSize, while U-Net improves \SkillCount\ but hurts \SkillSize\ and \SkillRotation.
Thus the regression supports the paper's qualitative claim that no single architecture or conditioning choice gives uniformly reliable extrapolation across these basic skills.
We note that the analysis uses only $103$ rows, each variant changes one design choice from the baseline (so interactions between choices are not observed), and the held-in $R^2$ values should not be read as performance estimates for unseen designs.

\end{document}